\documentclass[10pt,twocolumn,letterpaper]{article}

\usepackage[pagenumbers]{cvpr}

\usepackage{algorithm}
\usepackage[noend]{algpseudocode}

\definecolor{cvprblue}{rgb}{0.21,0.49,0.74}
\usepackage[pagebackref,breaklinks,colorlinks,allcolors=cvprblue]{hyperref}

\usepackage{cuted}

\title{RT$^2$C: Closing the Tactile Feedback Loop in Real-Time Chunking for Dexterous Manipulation}

\author{Xingyu Ji\textsuperscript{1,2,*}\quad
  Tutian Tang\textsuperscript{1,3,*}\quad
  Wanli Xing\textsuperscript{2}\quad
  Wensi Huang\textsuperscript{2,3,4}\\
  Cewu Lu\textsuperscript{1,\textdagger}\quad
  Kaifeng Zhang\textsuperscript{2,\textdagger}\quad
  Qiaojun Yu\textsuperscript{3,\textdagger}\\[1pt]
  \multicolumn{1}{@{}c@{}}{\small
    \textsuperscript{1}Shanghai Jiao Tong University\enspace
    \textsuperscript{2}Sharpa\enspace
    \textsuperscript{3}Shanghai Artificial Intelligence Laboratory\enspace
    \textsuperscript{4}University of Science and Technology of China}\\
  {\small\href{mailto:ji_xingyu@sjtu.edu.cn}
    {\normalcolor\fontencoding{T1}\ttfamily ji\_xingyu@sjtu.edu.cn}}\\[1pt]
  {\small Project Page: \url{https://jxyjxyw.github.io/RT2C}}
}
\AddToHookNext{env/abstract/before}{\begingroup
  \footnotetext[1]{Equal contribution.}\footnotetext[2]{Corresponding authors.}\endgroup
}

\begin{document}

\twocolumn[{
    \renewcommand\twocolumn[1][]{#1}
    \maketitle
\vspace{-0.5cm}
\includegraphics[width=\textwidth]{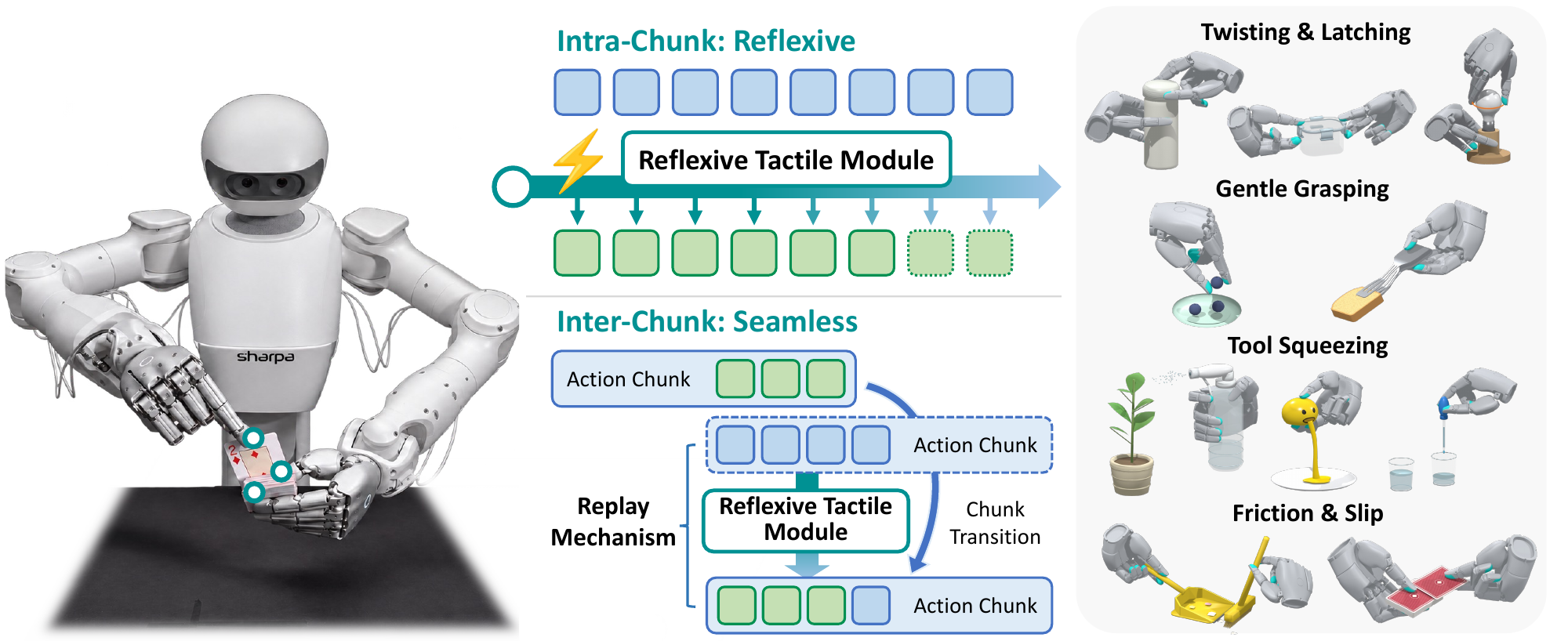}
\captionsetup{hypcap=false}
\captionof{figure}{RT$^2$C closes the tactile feedback loop within and across action chunks. Within each chunk, the \textit{Reflexive Tactile Module (RTM)} injects fresh tactile feedback into upcoming actions at every control step. Across chunks, the inference-time \textit{Feedback Replay} mechanism incorporates observations collected during planning into the incoming action sequence before chunk transition. The framework is evaluated on ten real-world dexterous manipulation tasks spanning four contact-interaction categories, illustrated on the right.}
\label{fig:teaser}
\captionsetup{hypcap=true}
\vspace{1cm}
}]

\begin{abstract}
Dexterous manipulation demands rapid, continuous adaptation to changing contact forces, yet current vision-language-action (VLA) models with chunked execution struggle to respond to tactile feedback within an action chunk.
When per-step tactile refinement is introduced into asynchronous real-time chunking (RTC), the executed actions can diverge from the fixed action prefix used to condition an incoming plan, risking discontinuities at chunk boundaries.
We present RT$^2$C, which equips a frozen backbone with a lightweight Reflexive Tactile Module (RTM) to close the tactile feedback loop within and across action chunks.
The backbone asynchronously produces clean action plans.
Then the RTM lightly re-noises each plan and, at every control step, denoises a sliding window of upcoming actions conditioned on cached visual-language features and the latest tactile and proprioceptive feedback.
When a new plan arrives, the Feedback Replay mechanism makes sure the incoming chunk inherits the tactile corrections it missed during backbone inference and takes over seamlessly.
Evaluation on ten real-world bimanual dexterous tasks proves that RT$^2$C achieves notable gains on contact-sensitive manipulation.
Ablations confirm the value of fresh tactile feedback and Feedback Replay, and the compatibility with other backbones.
\end{abstract}

\section{Introduction}
\label{sec:intro}
Humans perform dexterous manipulation every day, such as preventing fragile objects from slipping, extracting thin cards under changing friction, and aligning interlocking parts.
Such dexterity relies on coordinating voluntary motor control with rapid sensory feedback through spinal reflex pathways and supraspinal circuits~\cite{johansson2009coding,pruszynski2012optimal}.
While recent vision-language-action models (VLAs)~\cite{intelligence2025pi05,kim2024openvla,yu2025forcevla,li2026forcevla2,zhang2025ta,huang2025tactilevla} and world-action models (WAMs)~\cite{yuan2026fast,wang2026openwam,ye2026world,cai2026internw0,li2026egowam} enable task-directed manipulation by predicting multi-step action chunks, their inference latency and chunked execution create a gap between high-level planning and fast contact control.
Closing this gap requires two complementary capabilities: \textit{reflexive adaptation} to changing contact conditions within each action chunk, and \textit{seamless transitions} across chunk boundaries without losing tactile responsiveness.

Existing approaches address only parts of this problem.
Direct tactile VLAs read tactile states only once per chunk request, whereas slow--fast tactile policies such as AT-VLA and T-Rex~\cite{li2026atvla,niu2026trex} and rolling denoising methods~\cite{xiong2026react,park2026pir2} either require joint training with a specific backbone, refresh tactile conditioning at coarse sub-chunk intervals, or evolve actions without explicit guidance from a clean high-level plan.
Meanwhile, real-time chunking (RTC) methods promote smooth transitions by conditioning incoming chunks on actions committed during planning~\cite{black2025trainingrtc,liu2026learning,jiang2026futurertc}, but do not incorporate real-time tactile feedback during execution.
Crucially, when fast feedback updates ongoing actions during planning latency, an incoming plan conditioned on a fixed request-time action prefix becomes misaligned with the tactile corrections accumulated during its computation. Such misalignment disrupts motion continuity and tactile responsiveness at chunk transitions.

We introduce RT$^2$C, which extends \textbf{R}eal-\textbf{T}ime \textbf{C}hunking (RTC) with a Reflexive \textbf{T}actile Module (RTM) to close the tactile feedback loop both within and across action chunks (\cref{fig:teaser}).
A frozen, deliberative VLA backbone such as $\pi_0$~\cite{black2026pi0} asynchronously generates clean plans in the form of action chunks with training-time RTC and supplies cached visual-language features to the lightweight RTM.
Conditioned on the clean plan, cached context, and recent proprioceptive, torque, and fingertip tactile histories, the RTM refines upcoming actions at every control step through staircase re-noising and sliding-window denoising.
Therefore, each command incorporates the latest observation immediately before execution.
Before a newly computed clean plan takes over, inference-time Feedback Replay rolls the new sequence forward through the observations logged during the backbone inference, aligning its denoising schedule and tactile corrections with the transition step.
Together, these mechanisms achieve step-level reflexive execution within chunks and seamless transitions across chunks, while the backbone stays frozen and is never trained jointly with the RTM.

We evaluate RT$^2$C on ten real-world bimanual dexterous manipulation tasks spanning gentle grasping and holding, squeezing tools, friction and slip, and twisting and latching.
Across the ten tasks, RT$^2$C achieves an average success rate of 76.5\%, exceeding the $\pi_0$ baseline by 5.0 percentage points, with gains of 15, 20, and 20 percentage points on Blueberry, Bread, and Card, respectively.
Controlled ablations verify the necessity of live tactile feedback, explicit clean-plan conditioning, and Feedback Replay, which lowers boundary jump ratios by 10.9\% and 5.4\% and transition jerk by 6.8\% and 20.3\% on two representative tasks, Blueberry and Card, respectively.

Our contributions are threefold:
\begin{itemize}
    \item \textbf{Reflexive tactile feedback within action chunks.}
    We introduce a Reflexive Tactile Module (RTM) that uses staircase re-noising and sliding-window denoising around a frozen backbone's clean plan to generate reflexive actions in real time, injecting fresh tactile and torque feedback at every control step.

    \item \textbf{Seamless, feedback-aligned chunk transitions.}
    We combine RTC action-prefix conditioning with an inference-time Feedback Replay mechanism, which incorporates feedback logged during the backbone inference into the incoming chunk, enabling seamless chunk transitions without interrupting tactile feedback.

    \item \textbf{Real-world evaluation.}
    We validate RT$^2$C across 10 bimanual dexterous tasks, demonstrating consistent gains over frozen backbones ($\pi_0$, $\pi_{0.5}$, and DM0.5) and smoother chunk transitions in online execution.

\end{itemize}

\section{Related Work}
\label{sec:related_work}

\paragraph{Tactile Feedback for Dexterous Manipulation.}
Action-chunking policies~\cite{zhao2023act,chi2023diffusionpolicy}, vision-language-action (VLA) models~\cite{kim2024openvla,black2026pi0,intelligence2025pi05,dexmal2026dm05}, and world-action models (WAMs)~\cite{yuan2026fast,wang2026openwam,ye2026world} achieve strong long-horizon manipulation by predicting multi-step action sequences, yet executing an entire chunk open-loop leaves the robot unable to react to sudden contact changes.
To incorporate tactile sensing into foundation policies, one line of work fuses tactile or force measurements directly into VLA representations or mixture-of-experts action heads~\cite{huang2025tactilevla,yu2025forcevla,li2026forcevla2,zhang2026tacvla,tang2026modevla}.
Because these models read tactile input only once per chunk request, they still cannot adjust ongoing motion when slip or force changes occur mid-chunk.
A second line of work adopts slow--fast architectures that pair slower visual-language or diffusion planning with a higher-frequency tactile pathway~\cite{xue2025rdp,xue2026tdp,bi2025vlatouch,li2026atvla,niu2026trex,yu2026omnitactune}.
Existing designs couple the two pathways through latent action representations~\cite{xue2025rdp,xue2026tdp}, visual-language context features~\cite{li2026atvla}, residual controllers that refine base trajectories~\cite{bi2025vlatouch,yu2026omnitactune}, or cascaded tactile denoising~\cite{niu2026trex}.
However, these approaches either require joint end-to-end training with a specific backbone, learn local action residuals without generative horizon refinement, or refresh tactile conditioning only at sub-chunk intervals as in T-Rex~\cite{niu2026trex}, leaving multiple action steps between feedback updates.
In contrast, RT$^2$C conditions a lightweight Reflexive Tactile Module (RTM) on the clean plan and cached visual-language features of a frozen deliberative backbone, injecting tactile and torque feedback at every control step for \textit{reflexive} execution while adapting across diverse backbones.

\textbf{Reactive Action Generation.}
Standard diffusion and flow-matching policies~\cite{chi2023diffusionpolicy,black2026pi0} must complete all denoising steps before executing an action chunk, which limits their closed-loop reactivity.
Diffusion Forcing~\cite{chen2024diffusionforcing} addresses this bottleneck by assigning per-token noise levels along the prediction horizon, allowing imminent actions to reach zero noise for execution while future actions remain at higher noise levels to absorb subsequent information.
Recent robot control methods build on this principle to stream actions during ongoing execution~\cite{xiong2026react,park2026pir2,lu2026faster}.
Specifically, REACT~\cite{xiong2026react} decouples visual-language encoding from rolling action denoising so that front action blocks are executed while Gaussian noise is appended at the tail.
$\pi\mathrm{R}^2$~\cite{park2026pir2} aligns the rolling denoising schedule with model inference latency to incorporate fresh proprioceptive states alongside asynchronous visual-language features, and FASTER~\cite{lu2026faster} accelerates flow-based VLA sampling for real-time response.
Despite these advances, existing rolling policies evolve a single generation stream from Gaussian noise and typically refresh observations at block intervals rather than guiding step-level tactile updates with an explicit high-level plan.
Inspired by partial re-noising~\cite{meng2022sdedit}, RT$^2$C instead constructs a staircase-noised action sequence directly around the backbone's clean plan and advances a sliding denoising window at every control tick, enabling each executed command to accumulate multi-step tactile refinements under explicit clean-plan guidance.

\textbf{Asynchronous Planning and Execution.}
Chunked policies commonly mitigate boundary discontinuities and inference delays through receding-horizon execution~\cite{chi2023diffusionpolicy}, temporal ensembling of overlapping chunks~\cite{zhao2023act}, or latency-aware scheduling~\cite{lu2026faster,park2026pir2}.
To explicitly align successive chunks under asynchronous inference, Real-Time Chunking (RTC)~\cite{black2025rtc}, training-time RTC~\cite{black2025trainingrtc}, and recent continuation methods~\cite{liu2026learning,jiang2026futurertc} condition each incoming plan on the action prefix committed during the planning delay.
However, when high-frequency tactile feedback modifies ongoing actions during planning latency, an incoming plan conditioned only on a fixed request-time prefix becomes misaligned with the executed trajectory and lacks the tactile corrections accumulated during its computation.
RT$^2$C bridges this gap by combining RTC prefix conditioning with inference-time Feedback Replay, rolling the incoming action sequence forward through the observations logged during backbone inference to achieve \textit{seamless} chunk transitions.

\section{Method}
\label{sec:method}

\begin{figure*}
  \centering
  \includegraphics[width=\textwidth]{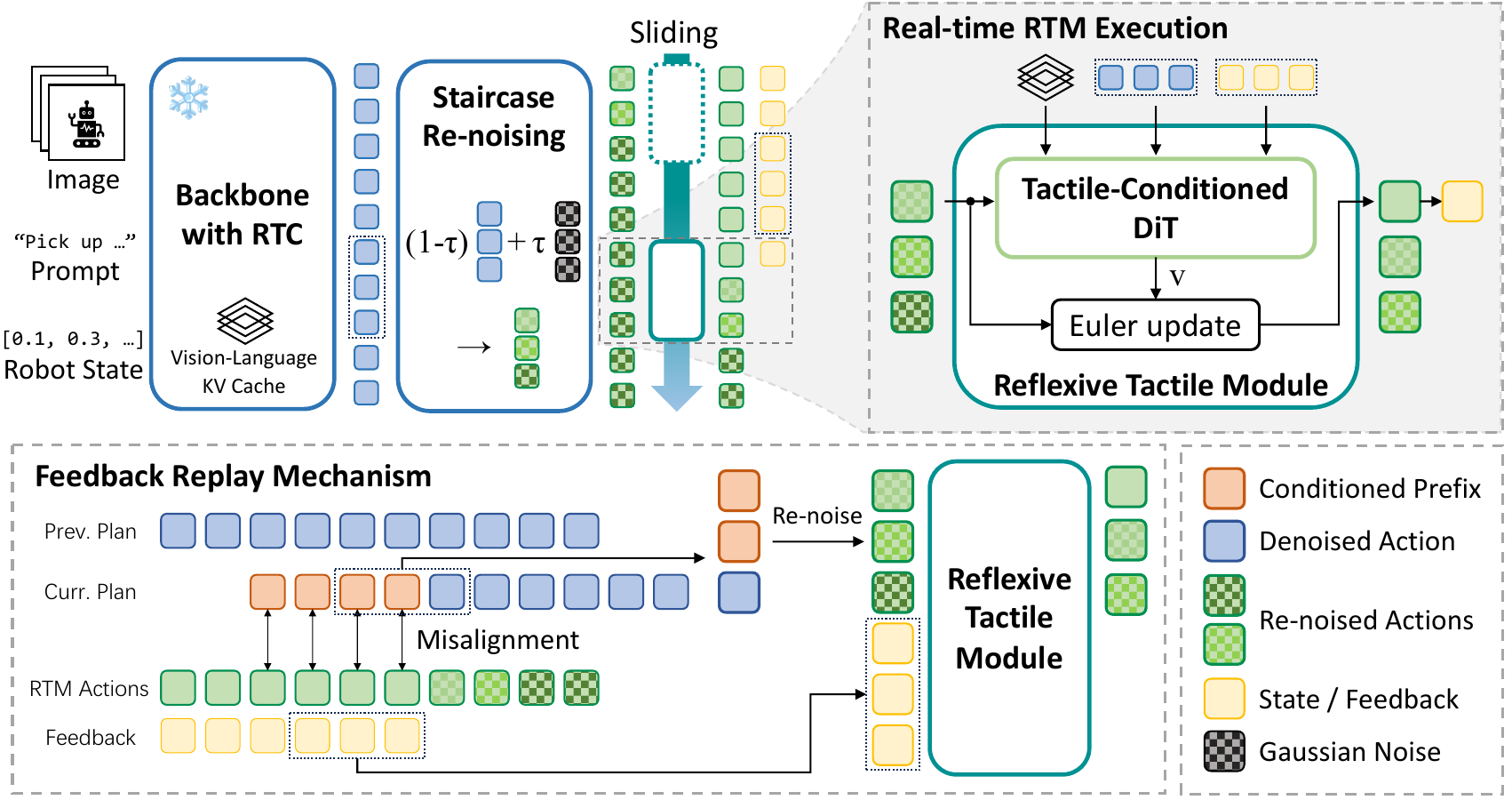}
  \caption{\textbf{Overview of RT$^2$C}. The backbone generates a clean action plan and caches visual-language keys and values. Staircase re-noising initializes the action sequence with lower noise for imminent actions. At each control step, the RTM uses the clean plan, cached context, and recent state, torque, and tactile feedback to predict velocities and denoise upcoming actions through an Euler update. The first action is executed, and the window advances. \textbf{Bottom:} Feedback Replay applies observations logged during planning to the incoming sequence, accounting for feedback that was unavailable when its RTC prefix was fixed.}
  \label{fig:method_overview}
\end{figure*}

Our method builds on a flow-based deliberative VLA backbone that produces clean plans at a low rate, paired with a Reflexive Tactile Module (RTM) that responds to the latest tactile observations (\cref{fig:method_overview}).
\Cref{sec:method_framework} formulates the problem.
\Cref{sec:method_fast} describes how the RTM injects tactile feedback into the backbone's clean plan through staircase re-noising and sliding-window denoising.
\Cref{sec:method_handoff} explains how the inference-time Feedback Replay mechanism achieves seamless chunk transitions.

\subsection{Overview}
\label{sec:method_framework}

The robot is controlled at discrete ticks $t$ with period $\Delta t$.
At each tick it receives an observation $\mathbf{o}_t=(\mathbf{s}_t,\mathbf{f}_t,\mathbf{u}_t)$ with joint state $\mathbf{s}_t$, arm joint torques $\mathbf{f}_t$, and fingertip tactile observations $\mathbf{u}_t$.
Camera images $\mathbf{I}_t$ and the instruction $\ell$ are available to the backbone $\pi_\theta^{\mathrm{slow}}$.
The policy outputs an absolute joint-position target $\mathbf{a}_t\in\mathbb{R}^{D}$ at every tick.
A backbone forward pass requested at tick $r$ spans $d$ ticks to produce a new $H$-step clean plan $\mathbf{A}_r$, which takes over at tick $r+d$ and is executed by the RTM $\pi_\phi^{\mathrm{fast}}$ for $N$ control ticks ($t\in [r+d, r+d+N-1]$, with $d+N\le H$) before switching to the next plan:
\begin{align}
  (\mathbf{A}_r,\mathcal{C}_r) &= \pi_\theta^{\mathrm{slow}}\big(\mathbf{I}_r,\ell,\mathbf{s}_r,\mathbf{a}^{\mathrm{prefix}}_{r}\big), \label{eq:slow}\\
  \mathbf{a}_t &= \pi_\phi^{\mathrm{fast}}\big(\mathbf{o}_{t-h+1:t}\mid\mathbf{A}_r,\mathcal{C}_r,\mathbf{s}_r\big). \label{eq:fast}
\end{align}
Here $\mathbf{A}_r\in\mathbb{R}^{H\times D}$ is the clean plan, $\mathcal{C}_r$ is the cached visual-language representation, $\mathbf{a}^{\mathrm{prefix}}_r\in\mathbb{R}^{d\times D}$ is the request-time action prefix from the previously active clean plan for the latency interval $[r,r+d-1]$, and $h$ is the feedback history length.
We build the backbone on $\pi_0$~\cite{black2026pi0} with training-time RTC~\cite{black2025trainingrtc}.
Given clean actions $\mathbf{a}$ and noise $\boldsymbol{\epsilon}\sim\mathcal{N}(\mathbf{0},\mathbf{I})$, flow matching regresses a velocity field on the interpolant
\begin{equation}
  \mathbf{x}^{\tau}=(1-\tau)\,\mathbf{a}+\tau\,\boldsymbol{\epsilon},\qquad
  \mathbf{v}^{\star}=\boldsymbol{\epsilon}-\mathbf{a},
  \label{eq:flow}
\end{equation}
and sampling integrates from $\tau=1$ to $\tau=0$.
The first $d$ action tokens are supplied clean at $\tau=0$ as the prefix $\mathbf{a}^{\mathrm{prefix}}_r$, and only the remaining tokens are noised and supervised.

The RTM refines the clean plan $\mathbf{A}_r$ through staircase re-noising and sliding-window denoising.
Staircase re-noising forms a noisy action sequence $\mathbf{X}_r\in\mathbb{R}^{H\times D}$, where each action's flow time $\tau_m$ denotes its noise level.
Imminent actions receive less noise, and later actions receive more, up to a fixed cap.
At tick $t$, the RTM processes the $W$-step window $\mathbf{X}_r[k{:}k{+}W]$, where $k=t-r$ is the plan-relative offset and $W$ is the window length.
Conditioned on recent feedback $\mathbf{o}_{t-h+1:t}$ and cached visual-language context $\mathcal{C}_r$, sliding-window denoising progressively refines near-term actions as the window advances.
Upcoming actions retain earlier corrections and refresh feedback before reaching $\tau=0$ for execution.

\subsection{Reflexive Tactile Module}
\label{sec:method_fast}

\paragraph{Tactile-Conditioned DiT.}
The Tactile-Conditioned Diffusion Transformer (DiT) $F_\phi$ maps a noisy $W$-step action window and its flow times to velocities $\hat{\mathbf{V}}\in\mathbb{R}^{W\times D}$ (\cref{fig:method_overview}).
At local tick $k=t-r$, action token $j\in\{0,\dots,W-1\}$ combines projections of $\mathbf{X}_r[k+j]$ and the aligned clean-plan action $\mathbf{A}_r[k+j]$ with a flow-time embedding of $\tau_j$.
We concatenate the action tokens, embedded feedback history $\mathbf{o}_{t-h+1:t}$, and an embedded request state $\mathbf{s}_r$.
An embedding of the local offset $k$ is added to every RTM input token.
Each layer of the DiT attends to the corresponding VLM layer’s cached keys and values in $\mathcal{C}_r$.

\paragraph{Staircase re-noising and sliding-window denoising.}
We initialize the noisy action sequence $\mathbf{X}_r$ by partially re-noising the clean plan~\cite{meng2022sdedit}:
\begin{equation}
  \mathbf{X}_r[m]=(1-\tau_m)\,\mathbf{A}_r[m]+\tau_m\,\boldsymbol{\epsilon}_{r+m}.
  \label{eq:init}
\end{equation}
Here $\tau_m=\min\{(m+1)\delta,K\delta\}$ for $m=0,\dots,H-1$ follows a staircase schedule~\cite{chen2024diffusionforcing}, where $\delta$ is the integration step size and $K\le W$ is the number of actions refined per tick.
Imminent actions start at lower noise levels, while later actions are capped at $K\delta<1$.

At each tick $t$, $F_\phi$ predicts velocities for the $W$-step window $\mathbf{X}_r[k{:}k{+}W]$ conditioned on the latest feedback.
The first $K$ actions are refined with one Euler step:
\begin{equation}
  \mathbf{X}_r[k+j]\leftarrow\mathbf{X}_r[k+j]-\delta\,\hat{\mathbf{v}}_j,\qquad j=0,\dots,K-1.
  \label{eq:step}
\end{equation}
Each update reduces the corresponding flow time by $\delta$, bringing the first action to $\tau=0$.
Advancing the window by one action carries the partially refined future actions into the next tick, so window position $j\in\{0,\dots,W-1\}$ maintains the invariant pre-update flow time $\tau_j=\min\{(j+1)\delta,K\delta\}$ at every offset $k$.

\textbf{Training.}
We train the backbone with training-time RTC as described in \cref{sec:method_framework}, then freeze it while optimizing the Tactile-Conditioned DiT $F_\phi$.
For each training sample, we select an eligible request tick $r$ and a local offset $k$ from a demonstration, with $t=r+k$.
For every training batch, the frozen backbone generates the clean plan $\mathbf{A}_r$ and visual-language cache $\mathcal{C}_r$ online from the request-time inputs using an empty RTC prefix ($d=0$), following the standard RTC training paradigm~\cite{black2025trainingrtc}.
For each action $j\in\{0,\dots,W-1\}$ in the training window, we form $\mathbf{x}_j=(1-\tau_j)\mathbf{a}^{\star}_{t+j}+\tau_j\boldsymbol{\epsilon}_{t+j}$ using the demonstrated action $\mathbf{a}^{\star}_{t+j}$, Gaussian noise, and the invariant window flow time $\tau_j$, paired with the aligned clean-plan action $\mathbf{A}_r[k+j]$.
The network also receives $\mathbf{o}_{t-h+1:t}$, $\mathbf{s}_r$, $k$, and $\mathcal{C}_r$, following the Tactile-Conditioned DiT formulation.
We use staircase flow matching loss to supervise the first $K$ actions:
\begin{equation}
  \mathcal{L}_{\mathrm{fast}}(\phi)=\frac{1}{KD}\,\mathbb{E}\!\left[\sum_{j=0}^{K-1}\left\|\hat{\mathbf{v}}_j-\big(\boldsymbol{\epsilon}_{t+j}-\mathbf{a}^{\star}_{t+j}\big)\right\|_2^2\right].
  \label{eq:loss}
\end{equation}
Under the cap $K\delta<1$, the re-noised clean plan in \cref{eq:init} lies near the noisy demonstration manifold~\cite{meng2022sdedit}, allowing $K$ Euler updates along the learned velocity field to steer $\mathbf{A}_r$ toward feedback-consistent actions.

\subsection{Seamless Transitions with Feedback Replay}
\label{sec:method_handoff}

The actions sent to the robot by the RTM in \cref{sec:method_fast} diverge from the prefix used to condition RTC during backbone planning.
At request tick $r$, the backbone is conditioned on a fixed prefix $\mathbf{a}^{\mathrm{prefix}}_r$ from the active clean plan, as observations $\mathbf{o}_{r:r+d-1}$ for the planning interval have not yet arrived.
Meanwhile, the RTM continues to refine outgoing actions using this newly arriving feedback, leaving the returned clean plan unaware of the tactile corrections accumulated during its computation.
Directly switching to this clean plan would discard their effect on upcoming actions.
To achieve seamless transitions, we introduce \textit{inference-time feedback replay} (Feedback Replay).

\begin{figure}
  \centering
  \includegraphics[width=\linewidth]{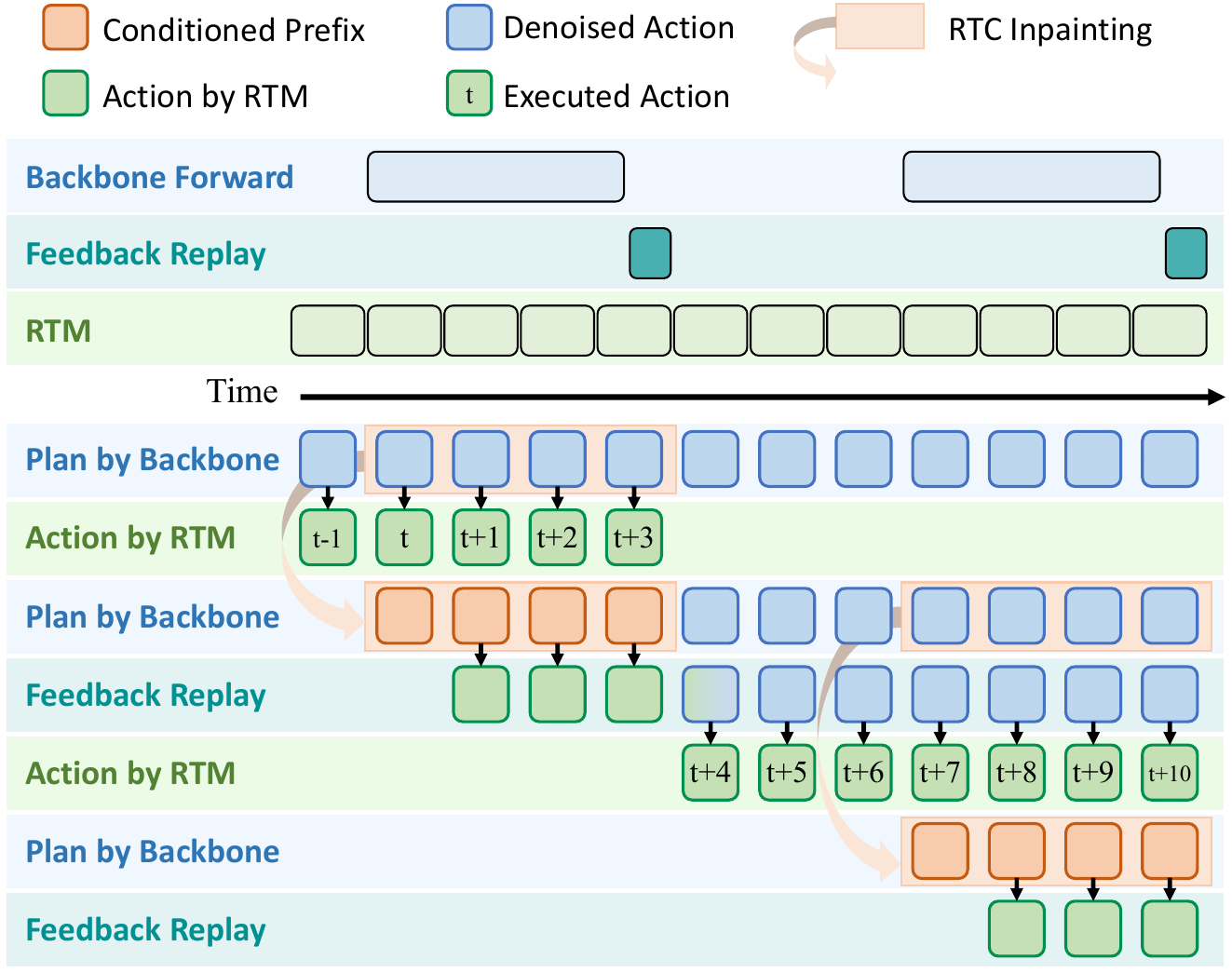}
  \caption{\textbf{Timeline of Execution.} \textbf{Above the time axis}, the computation schedule shows backbone inference, Feedback Replay, and per-tick RTM updates. The RTM continues refining the active plan while the backbone computes the next plan, followed by replay before takeover. \textbf{Below the time axis}, successive action sequences illustrate RTC prefix conditioning, feedback-driven RTM, and execution across chunk transitions.
  Orange, blue, and green blocks denote conditioned prefix actions, backbone-generated actions, and RTM-refined actions, respectively. Labeled green blocks ($t-1$, $t$, ..., $t+10$) indicate executed actions.}
  \label{fig:method_timeline}
\end{figure}

\paragraph{Feedback Replay.}
During planning, we log observations $\mathbf{o}_r,\dots,\mathbf{o}_{r+d-1}$ and retain the preceding $h-1$ observations needed to reconstruct each RTM input history.
Once the new clean plan $\mathbf{A}_r$ and cache $\mathcal{C}_r$ arrive, we initialize the noisy action sequence $\mathbf{X}_r$ by staircase re-noising in \cref{eq:init}.
Starting from the request-aligned beginning of the sequence, we replay offsets $k=0,\dots,d-1$ in chronological order.
At each offset, the RTM reads the logged history $\mathbf{o}_{r+k-h+1:r+k}$ and processes the window $\mathbf{X}_r[k{:}k{+}W]$, conditioned on the new clean plan, cache, and request state $\mathbf{s}_r$.
It applies \cref{eq:step} to the first $K$ actions and reduces their noise levels by $\delta$.
The window then advances by one action, carrying the updated future actions into the next replay step.
Repeating this update across $k=0,\dots,d-1$ advances the window to offset $d$ with front noise levels $\delta,2\delta,\dots,K\delta$, aligning the new sequence with the transition tick $t=r+d$.

At transition, the RTM uses the current observation history for one live update before sending the first command from the new sequence.
For $d\ge K-1$, this action has received $K-1$ replay updates at offsets $d-K+1,\dots,d-1$; the live update completes its $K$-step refinement with the latest feedback.
Subsequent actions continue along the same staircase schedule, combining earlier replay updates with new observations.
\Cref{alg:method_loop} summarizes this shared update procedure, with one RTM pass per live tick and $d$ additional passes per transition.
Since $F_\phi$ is lightweight and reuses $\mathcal{C}_r$, backbone inference and all $d$ replay passes complete within the $d$-tick planning budget (\cref{sec:exp_setup}).

\paragraph{Execution timeline.}
Following \cref{alg:method_loop}, \cref{fig:method_timeline} illustrates a request launched at tick $t$ while the RTM continues executing the active plan through $t+3$.
Once the new clean plan returns, the Feedback Replay processes the logged observations in order, advancing the incoming window while discarding replay outputs.
At $t+4$, a live RTM update incorporates the latest feedback before the first action from the new sequence is sent to the robot.
The same planning--replay--execution cycle repeats at subsequent transitions.

\begin{algorithm}[t]
  \caption{Online tactile control with Feedback Replay}
  \label{alg:method_loop}
  \small
  \begin{algorithmic}[1]
    \Function{RTMStep}{$\mathbf{X},k,q$}
      \State Read feedback $\mathbf{o}_{q-h+1:q}$ and plan context
      \State Predict velocities on window $k{:}k{+}W$ using $F_\phi$
      \State Update actions $k,\dots,k{+}K{-}1$ by \cref{eq:step}
      \State Decrease their noise levels by $\delta$
      \State \Return $\mathbf{X}[k]$
    \EndFunction
    \State Obtain a startup clean plan with an empty RTC prefix
    \State Initialize $\mathbf{X}$ by \cref{eq:init}; set $r\gets0$, $t\gets0$
    \For{each control tick $t$}
      \State Read $\mathbf{o}_t$ and append it to the log
      \If{request $r'$ arrives for takeover at $t=r'+d$}
        \State Initialize $\mathbf{X}'$ from $\mathbf{A}_{r'}$ by \cref{eq:init}
        \For{$k=0,\dots,d-1$} \Comment{Feedback Replay}
          \State \Call{RTMStep}{$\mathbf{X}',k,r'+k$} \Comment{discard output}
        \EndFor
        \State Switch to $\mathbf{X}'$ and its plan context; $r\gets r'$
      \EndIf
      \If{a new request is due}
        \State $r'\gets t$; $\mathbf{a}^{\mathrm{prefix}}_{r'}\gets\mathbf{A}_r[r'{-}r{:}r'{-}r{+}d]$
        \State Launch $\pi^{\mathrm{slow}}$ with $\mathbf{I}_{r'}$, $\ell$, $\mathbf{s}_{r'}$, and prefix
      \EndIf
      \State $\mathbf{a}_t\gets$ \Call{RTMStep}{$\mathbf{X},t-r,t$}
      \State Send $\mathbf{a}_t$ to the robot
    \EndFor
  \end{algorithmic}
\end{algorithm}

\section{Experiments}
\label{sec:exp}

We evaluate the framework on 10 real-world dexterous tasks and organize the experiments around 4 questions.\\
\textbf{Q1}: How does RT$^2$C compare with baselines (\cref{sec:exp_main})?\\
\textbf{Q2}: Which inputs of the RTM are necessary (\cref{sec:exp_ablation})?\\
\textbf{Q3}: Does Feedback Replay achieve smooth action chunk transitions (\cref{sec:exp_replay})?\\
\textbf{Q4}: Does the RTM adapt to other backbones (\cref{sec:exp_plugin})?

\subsection{Experimental Setup}
\label{sec:exp_setup}

\paragraph{Hardware.}
All experiments use a Sharpa North bimanual robot with two 7-DoF arms, two 22-DoF Sharpa W01 dexterous hands with fingertip tactile sensors, and a 5-DoF waist.
Visual observations are captured by one head-mounted camera and two wrist-mounted cameras, one on each wrist.
Each fingertip integrates a vision-based tactile sensor that provides six-dimensional force/torque measurements, yielding a 60-dimensional tactile input across all ten fingertips.
Each arm additionally reports seven joint torques.
Commands are sent at 30\,Hz.
The VLA backbone and the RTM run on a single NVIDIA GeForce RTX 4090.

\paragraph{Implementation details.}
We use $\pi_0$ with training-time RTC as the backbone. We set $H=50$, $d=4$ and $N=20$.
A fraction (20\%) of training samples use $d=0$, so the same model also produces the initial clean plan without a prefix.
The backbone is trained for 100K steps on 8 NVIDIA A800 GPUs with the AdamW optimizer with cosine learning rate decay and color jitter augmentation on input images.
The denoising step is set to 10.

For the RTM, we use $h=4$, $W=24$, $K=3$, and $\delta=0.1$, and sample plan offsets uniformly from $k\in\{0,\dots,23\}$ during training.
The DiT in the RTM is randomly initialized and has 9 transformer layers with approximately 160M trainable parameters.
We train the RTM for 50K steps.
The whole training pipeline finishes within 24 hours.
As for inference speed, the backbone planning and a single live RTM update take 84.4\,ms and 2.8\,ms on average, respectively.
The summed runtime of backbone planning, Feedback Replay updates, and one live RTM update is 93.3\,ms on average, which is elegantly within the $d=4$ control-step budget (133.3\,ms at 30\,Hz).

\begin{figure*}
  \centering
  \includegraphics[width=\textwidth]{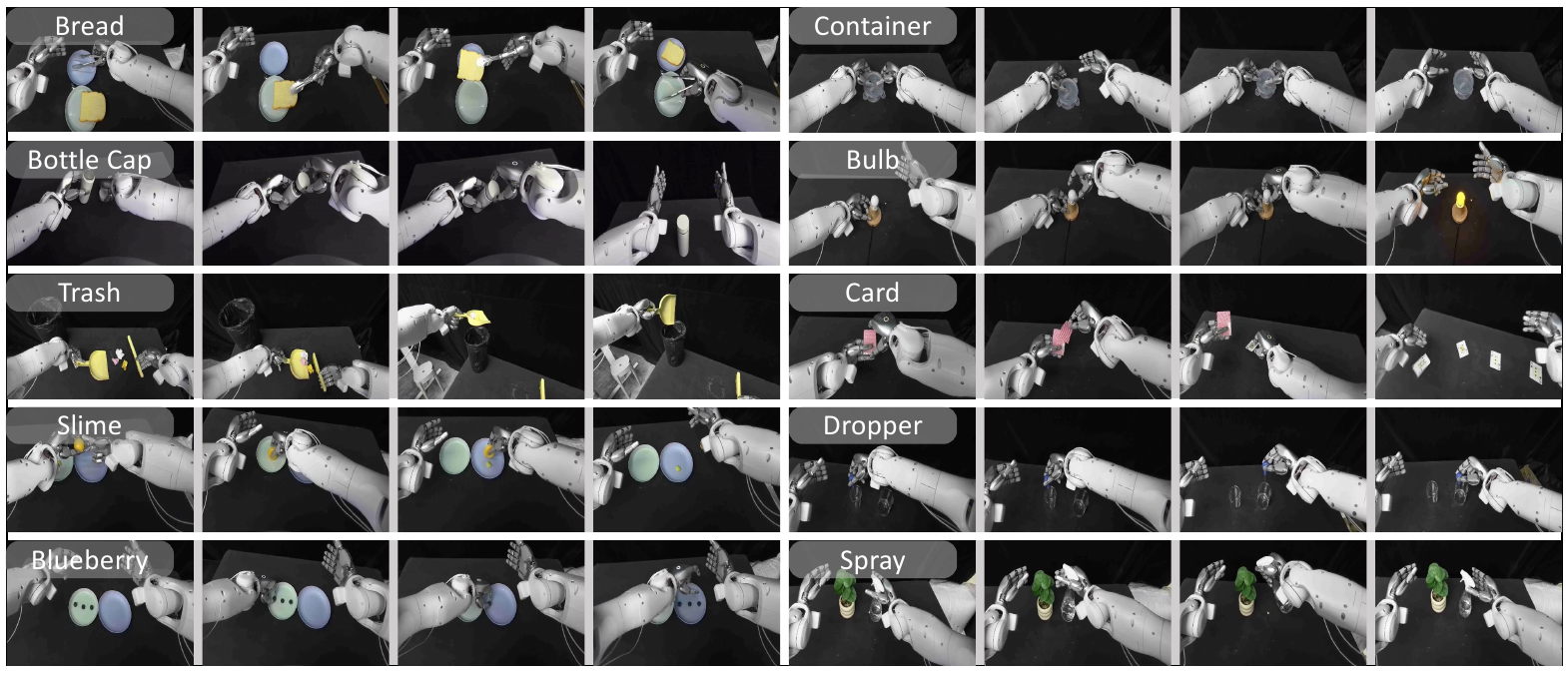}
  \caption{Ten dexterous manipulation tasks, illustrated with four keyframes from one teleoperated demonstration per task. Left, top to bottom: Bread, Bottle cap, Trash, Slime, and Blueberry. Right: Container, Bulb, Card, Dropper, and Spray. Within each sequence, frames progress from left to right. The task groups are gentle grasping and holding (Blueberry, Bread), tool squeezing (Spray, Slime, Dropper), friction and slip (Card, Trash), and twisting and latching (Bottle cap, Bulb, Container).}
  \label{fig:exp_tasks}
\end{figure*}

\paragraph{Tasks.}
We evaluate 10 tasks grouped into four categories by the role of touch (\cref{fig:exp_tasks}).
\textbf{Gentle grasping and holding}: moving blueberries without crushing them (Blueberry) and moving a slice of bread between plates with tongs (Bread).
\textbf{Tool Squeezing}: picking up a spray bottle, spraying a plant (Spray), transferring slime between plates with a suction toy (Slime), and transferring liquid between cups with a dropper (Dropper).
\textbf{Friction and slip}: dealing cards from a stack (Card) and sweeping the trash and dumping it into a trash can (Trash).
\textbf{Twisting and latching}: screwing on a bottle cap (Bottle cap), tightening a light bulb until it lights up (Bulb), and closing a food container and fastening all clips (Container).
We collect roughly 250 teleoperated demonstrations per task using Sharpa AE01 data gloves, which provide vibrotactile feedback to the operator.

\subsection{Comparison with Baselines}
\label{sec:exp_main}

We compare our method with five baselines: vision-only \textbf{$\pi_0$}~\cite{black2026pi0}, \textbf{$\pi_{0.5}$}~\cite{intelligence2025pi05}, and \textbf{DM0.5}~\cite{dexmal2026dm05}; tactile \textbf{MoDE-VLA}~\cite{tang2026modevla}; and streaming vision-only \textbf{REACT}~\cite{xiong2026react}.
All baselines are trained on the same demonstrations.

\begin{table*}[t]
  \centering
  \caption{Comparison with baselines on ten tasks in four groups. Success rate (\%) over 20 trials per task for all methods.}
  \label{tab:exp_main}
  \small
  \setlength{\tabcolsep}{4pt}
  \begin{tabular}{@{}l*{10}{c}@{}}
    \toprule
    & \multicolumn{2}{c}{\shortstack{Gentle grasping\\and holding}} & \multicolumn{3}{c}{Tool squeezing} & \multicolumn{2}{c}{\shortstack{Friction\\and slip}} & \multicolumn{3}{c}{\shortstack{Twisting\\and latching}} \\
    \cmidrule(lr){2-3}\cmidrule(lr){4-6}\cmidrule(lr){7-8}\cmidrule(lr){9-11}
    Method & Blueberry & Bread & Spray & Slime & Dropper & Card & Trash & Bottle cap & Bulb & Container \\
    \midrule
    $\pi_0$~\cite{black2026pi0} & 80 & 75 & \textbf{100} & 30 & \textbf{100} & 55 & 75 & 45 & 85 & 70 \\
    $\pi_{0.5}$~\cite{intelligence2025pi05} & 10 & 90 & \textbf{100} & 10 & 90 & 10 & 80 & 35 & 70 & \textbf{85} \\
    DM0.5~\cite{dexmal2026dm05} & 45 & 65 & \textbf{100} & 10 & 25 & 10 & 10 & 15 & 45 & 80 \\
    MoDE-VLA~\cite{tang2026modevla} & 80 & \textbf{95} & \textbf{100} & \textbf{45} & 95 & \textbf{75} & 60 & \textbf{60} & 80 & 75 \\
    REACT~\cite{xiong2026react} & 30 & 30 & \textbf{100} & 30 & 10 & 50 & \textbf{90} & 40 & 50 & 65 \\
    \midrule
    Ours & \textbf{95} & \textbf{95} & \textbf{100} & 30 & \textbf{100} & \textbf{75} & 75 & 30 & \textbf{90} & 75 \\
    \bottomrule
  \end{tabular}
\end{table*}

\Cref{tab:exp_main} reports the results.
On gentle grasping and holding, our method reaches 95\% on both tasks, improving over $\pi_0$ by 15 and 20 percentage points on Blueberry and Bread, respectively.
It also outperforms MoDE-VLA on Blueberry and matches it on Bread.
On Card, our method improves over $\pi_0$ from 55\% to 75\%, matching MoDE-VLA, while on Bulb it reaches 90\%, exceeding $\pi_0$ and MoDE-VLA by 5 and 10 percentage points, respectively.
On Slime, however, our method fails to achieve good results.
A typical failure mode is getting stuck at one location, where the policy repeatedly repositions and squeezes the slime but cannot generate sufficient suction to lift it.
We suspect this is because the task has relatively high tolerance and does not require fine tactile feedback, so the RTM's per-step corrections can become counterproductive.
In contrast, on Dropper, MoDE-VLA and $\pi_0$ often draw too much liquid, whereas our method consistently extracts an appropriate amount, suggesting that per-step tactile feedback yields clear benefits for tasks that require precise contact dynamics.
Overall, the comparison with MoDE-VLA highlights task-dependent gains instead of a uniform advantage with per-step tactile feedback.
Future work may further explore how to make tactile feedback on-demand.
Since these methods differ in more than their tactile update frequency, we examine the contribution of fresh tactile input separately in \cref{sec:exp_ablation}.

\subsection{Ablation of the Reflexive Tactile Module}
\label{sec:exp_ablation}

We ablate the inputs of the RTM on Blueberry, Card, and Bulb, covering three task groups (\cref{tab:exp_ablation}).
\emph{w/o tactile \& torque} retrains the RTM without fingertip tactile and arm-torque inputs.
\emph{Frozen request tactile} retrains the RTM with all tactile history slots set to the frame captured at the backbone request tick. At inference, this frame stays fixed throughout Feedback Replay and live execution of that clean plan; state and torque retain their live four-frame histories.
\emph{w/o VLM KV cache} retrains it without the backbone's prefix cache, so it sees no visual-language context.
\emph{w/o explicit plan} retrains it without the explicit clean-plan projection and conditioning branch. VLM KV and feedback remain available, and inference retains the same initialization by re-noising the backbone's clean plan. This variant isolates the contribution of explicit clean-plan conditioning.

\begin{table}[t]
  \centering
  \caption{Ablation of RTM inputs on Blueberry, Card, and Bulb. We report success rates (\%) over 20 trials per task. Avg. denotes the average success rate.}
  \label{tab:exp_ablation}
  \small
  \setlength{\tabcolsep}{3pt}
  \renewcommand{\arraystretch}{1.15}
  \begin{tabular}{lcccc}
    \toprule
    Variant & Blueberry & Card & Bulb & Avg. \\
    \midrule
    w/o tactile \& torque & 60 & 0 & 80 & 47 \\
    Frozen request tactile & 65 & 45 & 90 & 67\\
    w/o VLM KV cache & 35 & 15 & \textbf{100} & 50\\
    w/o explicit plan & 0 & 5 & 15 & 6.7 \\
    Ours (full) & \textbf{95} & \textbf{75} & 90 & \textbf{87} \\
    \bottomrule
  \end{tabular}
\end{table}

\paragraph{Tactile feedback.}
\Cref{tab:exp_ablation} shows that removing tactile and torque inputs reduces success on all three tasks, with the largest drop on Card, from 75\% to 0\%.
The average success rate falls from 86.7\% to 46.7\%.
Freezing touch at the backbone request reduces success by 30 percentage points on both Blueberry and Card, while Bulb remains at 90\%.
The average falls to 66.7\%.
The comparison supports the value of fresh tactile feedback on Blueberry and Card, beyond the tactile observation available at the request tick.

\paragraph{Backbone context and explicit clean-plan conditioning.}
Removing the VLM KV cache reduces success to 35\% on Blueberry and 15\% on Card, but increases it from 90\% to 100\% on Bulb.
Its average success rate is 50.0\%.
The benefit of visual-language context is therefore task-dependent in this evaluation.
Removing explicit clean-plan conditioning yields low success across all three tasks: 0\%, 5\%, and 15\% on Blueberry, Card, and Bulb, respectively.
The average falls to 6.7\%.
These results support explicitly conditioning refinement on the clean plan beyond using the clean plan only for initialization.

\subsection{Smoothness at Action Chunk Transitions}
\label{sec:exp_replay}

To answer \textbf{Q3},
we compare three execution modes on Blueberry and Card. For each task, we select 5 successful rollouts to evaluate.
\textbf{RTC async} continues execution during planning.
\textbf{RTM w/o replay} initializes the incoming sequence at chunk transitions.
\textbf{RTM + replay} first processes the feedback recorded during planning via Feedback Replay.
All these experiments share the same checkpoints and controller configurations.

We extract arm and hand positions $\mathbf{x}_i\in\mathbb{R}^{58}$ at timestamps $t_i$ from the recorded execution log.
Let $\mathcal{B}$ contain the first received sample at each chunk transition, and let $\mathcal{I}$ contain within-chunk differences.
For $\Delta\mathbf{x}_i=\mathbf{x}_i-\mathbf{x}_{i-1}$, the jump ratio is
\begin{equation}
  J=\frac{|\mathcal{B}|^{-1}\sum_{b\in\mathcal{B}}\|\Delta\mathbf{x}_b\|_2}{|\mathcal{I}|^{-1}\sum_{i\in\mathcal{I}}\|\Delta\mathbf{x}_i\|_2}.
  \label{eq:jump}
\end{equation}
We use divided differences $\mathbf{D}^{(0)}_i=\mathbf{x}_i$ and
\begin{equation}
  \mathbf{D}^{(k)}_i=k\frac{\mathbf{D}^{(k-1)}_{i+1}-\mathbf{D}^{(k-1)}_i}{t_{i+k}-t_i},\quad k=1,2,3.
  \label{eq:online_difference}
\end{equation}
Then, transition jerk RMS uses the three four-point stencils crossing each boundary:
\begin{equation}
  R_{\mathrm{t}}=\left(\frac{1}{3|\mathcal{B}|}\sum_{b\in\mathcal{B}}\sum_{s=1}^{3}\|\mathbf{D}^{(3)}_{b-s}\|_2^2\right)^{1/2}.
  \label{eq:online_jerk}
\end{equation}
\begin{table}
  \centering
  \caption{Online smoothness from raw measured state. $J$: jump ratio; $R_{\mathrm{t}}$: transition jerk RMS in $10^3$ rad/s$^3$. Values are arithmetic means of per-episode metrics, with equal episode weight. Green superscripts show relative improvement over RTM w/o replay.}
  \label{tab:exp_replay}
  \small
  \setlength{\tabcolsep}{1.5pt}
  \begin{tabular}{@{}lr@{}lr@{}lr@{}lr@{}l@{}}
\toprule
& \multicolumn{4}{c}{Blueberry} & \multicolumn{4}{c}{Card} \\
\cmidrule(lr){2-5}\cmidrule(l){6-9}
Execution & \multicolumn{1}{c}{$J\downarrow$} & & \multicolumn{1}{c}{$R_{\mathrm{t}}\downarrow$} & & \multicolumn{1}{c}{$J\downarrow$} & & \multicolumn{1}{c}{$R_{\mathrm{t}}\downarrow$} & \\
\midrule
RTC async & 1.090 &  & 0.736 &  & 1.103 &  & 0.493 &  \\
RTM w/o replay & 1.246 &  & 0.601 &  & 1.187 &  & 0.680 &  \\
RTM + replay & 1.111 & \textsuperscript{\textcolor[rgb]{0,0.45,0.20}{+10.9\%}} & 0.560 & \textsuperscript{\textcolor[rgb]{0,0.45,0.20}{+6.8\%}} & 1.123 & \textsuperscript{\textcolor[rgb]{0,0.45,0.20}{+5.4\%}} & 0.542 & \textsuperscript{\textcolor[rgb]{0,0.45,0.20}{+20.3\%}} \\
\bottomrule
\end{tabular}

\end{table}
With Feedback Replay, mean per-episode $J$ is 10.9\% lower on Blueberry and 5.4\% lower on Card (\cref{tab:exp_replay}).
Mean per-episode transition jerk RMS decreases by 6.75\% on Blueberry and by 20.3\% on Card.

\subsection{Ablation on Backbones}
\label{sec:exp_plugin}

\begin{table}
  \centering
  \caption{Backbone ablations on Blueberry with 20 trials. Base: the original backbone results. +RTC: the backbone executing its own clean plans with RTC. +RTC+RTM: the same frozen RTC backbone with our RTM; for $\pi_0$ this is our full method. $\Delta$ denotes the gain of +RTC+RTM over +RTC in percentage points.}
  \label{tab:exp_plugin}
  \small
  \setlength{\tabcolsep}{4pt}
  \renewcommand{\arraystretch}{1.15}
  \begin{tabular}{@{}lcccc@{}}
    \toprule
    Backbone & Base & +RTC & +RTC+RTM & $\Delta$ (pp) \\
    \midrule
    $\pi_0$ & 80 & 75 & 95 & +20 \\
    $\pi_{0.5}$ & 10 & 25 & 35 & +10 \\
    DM0.5 & 45 & 35 & 40 & +5 \\
    \bottomrule
  \end{tabular}
\end{table}

We attach the RTM to three backbones on the Blueberry task, $\pi_0$~\cite{black2026pi0}, $\pi_{0.5}$~\cite{intelligence2025pi05}, and DM0.5~\cite{dexmal2026dm05} (\cref{tab:exp_plugin}).
Each backbone is fine-tuned on the same demonstrations, then frozen while a separate RTM is trained for 50K steps.
For $\pi_0$ and $\pi_{0.5}$, we use identical RTM settings.
For DM0.5, we use cross-attention to read its native KV caches and select 8 VLM layers, each paired with one RTM layer, to keep the RTM parameter count comparable.
The resulting RTMs contain 159.6M parameters for $\pi_0$/$\pi_{0.5}$ and 159.1M for DM0.5.
The feedback history, action window, and denoising settings remain the same (\cref{sec:exp_setup}).
\Cref{tab:exp_plugin} shows the RTM raises $\pi_{0.5}$ from 25\% to 35\% and DM0.5 from 35\% to 40\%. It proves that the proposed method can be adapted to different backbones with action chunking.

\section{Conclusion}
\label{sec:conclusion}

We present RT$^2$C, a framework for reflexive and seamless dexterous manipulation that closes the tactile feedback loop both within and across action chunks.
Built on an asynchronous backbone with RTC, a separate Reflexive Tactile Module uses cached visual-language context and the latest tactile and proprioceptive feedback to update upcoming actions at every control step.
Staircase re-noising and sliding-window denoising accumulate successive feedback-conditioned updates before execution, while the inference-time Feedback Replay mechanism processes observations collected during backbone planning into the incoming action sequence before takeover.
Together, these mechanisms maintain tactile feedback through chunk transitions without retraining the frozen backbone.
Our evaluation spans ten real-world bimanual dexterous tasks, with controlled ablations examining the roles of live tactile feedback and Feedback Replay in manipulation reliability and motion continuity.
RT$^2$C offers a way to connect task-level planning with continuous contact-driven action correction.

\section*{LLM Usage Statement}
OpenAI Codex, Google Gemini, and DeepSeek were used to assist with code implementation, manuscript review, and manuscript drafting, including feedback on methodological exposition, experimental design, and related-work positioning. The authors retain full responsibility for the accuracy, originality, and integrity of the paper, including all scientific claims, experimental results, and references.

{
    \small
    \bibliographystyle{ieeenat_fullname}
    \bibliography{main}
}

\clearpage
\appendix
\counterwithin{equation}{section}
\twocolumn[{\pdfbookmark[0]{Appendix}{appendix-start}\centering\Large\bfseries Appendix\par\bigskip
}]

\section{Implementation Details}
\label{appendix:implementation}

\subsection{Network Architecture}
The default $\pi_0$ backbone takes images from the head and both wrist cameras, the language instruction, and the robot state. At inference, the RTC prefix is empty at startup. Later requests use the four actions from the active clean plan that cover the planning interval.
The Reflexive Tactile Module (RTM) uses the backbone's 50-step plan, its visual-language key--value (KV) caches, and the latest feedback to generate reflexive and seamless actions at 30\,Hz.

At each control tick, the RTM takes the four most recent frames of proprioceptive and tactile feedback as input. In each frame, the 63-dimensional joint state maps to one token, the 14 joint torques to two tokens (one per arm), and the 60 fingertip force and torque readings to ten tokens (one per fingertip). We add two learned embeddings to each of these 52 feedback tokens to identify its sensor group and its position in the four-frame history.

At each of the $W=24$ window positions, we sum separate projections of the noisy and clean actions with a flow-time embedding to form one action token. We concatenate the 52 feedback tokens, one request-time joint-state token, and the 24 action tokens.
All 77 tokens share an embedding of the window's starting index $k$. For a request at tick $r$, the feedback history ends at $r+k$. During replay, $k$ therefore measures cache age at the recorded tick.

The RTM has 9 transformer layers with width 1024 and feed-forward width 4096, totaling 159.6M parameters. Attention uses 8 query heads and one KV head with head dimension 256. Each RTM layer reads the cache of its corresponding selected VLM layer. The $\pi_{0.5}$ variant uses the same RTM configuration. For DM0.5, we use 8 layers with cross-attention to its native KV caches, giving 159.1M parameters. All three variants share the feedback history, action window, and denoising settings.

\subsection{Network Training}
We first fine-tune the $\pi_0$ backbone for 100K steps following the standard training-time RTC pipeline.
In 80\% of samples, the first four actions form a clean prefix. The remaining 20\% use an empty prefix for episode startup. Prefix actions remain at zero flow time and are excluded from the prediction loss.

We then freeze the backbone and train the RTM for 50K steps. For every batch, the backbone generates a clean action plan and VLM caches online using an empty prefix. Demonstrations are split by episode, and request offsets are sampled uniformly from 0 to 23 so that training covers both replay and live execution. The detailed hyperparameters are listed in~\cref{appendix:tab:hyperparameters}.

The RTM operates in normalized absolute joint coordinates. The backbone outputs are first decoded to physical targets and then normalized with the RTM action statistics. State, torque, and tactile inputs use their own statistics.

For flow matching, Gaussian noise is mixed with the demonstrated action, while the predicted clean plan is supplied as a separate condition. The flow times across the window are $0.1,0.2,0.3,0.3,\ldots,0.3$.
The loss in \cref{eq:loss} supervises the first 3 action tokens.

\begin{table}[t]
\centering
\caption{\textbf{RTM training and inference hyperparameters}.}
\label{appendix:tab:hyperparameters}
\small
\begin{tabular*}{\columnwidth}{@{\extracolsep{\fill}}lr@{}}
\toprule
\textbf{Setting} & \textbf{Value} \\
\midrule
RTM training steps & 50K \\
Global batch size & 32 \\
Optimizer & AdamW \\
Peak learning rate & $2.5\times10^{-5}$ \\
Scheduled final learning rate & $2.5\times10^{-6}$ \\
Learning rate warmup & 1K steps \\
Cosine schedule length & 100K steps \\
AdamW first-moment decay $\beta_1$ & 0.9 \\
AdamW second-moment decay $\beta_2$ & 0.95 \\
AdamW stability constant $\epsilon$ & $10^{-8}$ \\
Weight decay & $10^{-10}$ \\
Global gradient norm clip & 1.0 \\
Compute precision & bfloat16 \\
\midrule
Backbone plan horizon $H$ & 50 actions \\
RTC prefix length $d$ (after startup) & 4 actions \\
Live actions per plan $N$ & 20 actions \\
Feedback history length $h$ & 4 frames \\
Action window length $W$ & 24 actions \\
Actions refined per tick $K$ & 3 actions \\
Euler step size $\delta$ & 0.1 \\
\bottomrule
\end{tabular*}
\end{table}

\begin{table*}[t]
\centering
\caption{\textbf{Task procedures and success criteria.} Tasks follow the order and grouping in \cref{tab:exp_main}.}
\label{appendix:tab:task_protocols}
\renewcommand{\arraystretch}{1.04}
\newcommand{\appendixTaskRow}[3]{\textbf{#1} & \raggedright #2 & \raggedright #3\tabularnewline[3pt]}
\newcommand{\appendixTaskDivider}{\noalign{\begingroup\color{black!15}\hrule height 0.25pt\endgroup\vskip 2pt}}
\begin{tabular}{@{}p{0.115\textwidth}@{\hspace{10pt}}p{0.49\textwidth}@{\hspace{12pt}}p{0.345\textwidth}@{}}
\toprule
\textbf{Task} & \textbf{Operation details} & \textbf{Success criteria} \\
\midrule
\appendixTaskRow{Blueberry}
  {Three blueberries start in a horizontal row on the green plate. With the left hand, pinch and lift each berry from left to right and place it in the matching position on the blue plate to the right.}
  {All three berries reach their matching positions without being dropped or damaged.}
\appendixTaskDivider
\appendixTaskRow{Bread}
  {Take the tongs from the blue plate with the right hand. Use them to move the bread slice from the green plate to the center of the blue plate, then set the tongs on the green plate.}
  {The bread is in the center of the blue plate and the tongs are on the green plate.}
\midrule[0.3pt]
\appendixTaskRow{Spray}
  {Pick up the spray bottle, aim it at the plant, and pull the trigger.}
  {The robot lifts the bottle and pulls the trigger without dropping the bottle.}
\appendixTaskDivider
\appendixTaskRow{Slime}
  {Pick up the suction toy from the blue plate on the right. Draw up the slime from the green plate on the left, move the toy over the blue plate, and squeeze the slime out.}
  {The toy picks up all the slime and deposits it on the blue plate without dropping any in transit.}
\appendixTaskDivider
\appendixTaskRow{Dropper}
  {Pick up the bulb dropper from the left beaker, draw liquid from the right beaker, and dispense it into the left beaker.}
  {Liquid drawn from the right beaker is delivered into the left beaker.}
\midrule[0.3pt]
\appendixTaskRow{Card}
  {Pick up the card case from the table with the right hand and pass it to the left. Use the right thumb and index finger to slide out three cards, one at a time, and deal them face up on the table.}
  {The left hand holds the case while the right hand deals one card at a time. All three cards land face up.}
\appendixTaskDivider
\appendixTaskRow{Trash}
  {Pick up the dustpan with the left hand and the broom with the right. Sweep the debris from the table into the pan, then use the left hand to empty the pan into the bin.}
  {Neither tool is dropped. The tabletop is clear and all the debris is in the bin.}
\midrule[0.3pt]
\appendixTaskRow{Bottle cap}
  {Pick up the bottle with the left hand. With the right hand, pick up the cap, place it on the bottle, and turn it until tight.}
  {The cap is threaded on straight and stays attached when the bottle is turned upside down.}
\appendixTaskDivider
\appendixTaskRow{Bulb}
  {Steady the base with the left hand and screw in the bulb with the right hand until it lights up.}
  {The bulb lights up.}
\appendixTaskDivider
\appendixTaskRow{Container}
  {Steady the container with the left hand and fasten the two clips on the right with the right hand. Then steady it with the right hand and fasten the other two clips with the left.}
  {All four clips are fastened.}
\bottomrule
\end{tabular}
\end{table*}

\subsection{Execution and Feedback Replay}
The backbone and RTM share one RTX 4090 GPU.
The backbone uses 10 denoising steps to generate an action plan.
We request the next plan four control ticks before each chunk transition and execute 20 live actions per plan. We retain the three observations before request tick $r$ and record the next four starting at $r$.
When the plan arrives, Feedback Replay processes offsets $k=0,1,2,3$ in order. Each pass uses the four-frame history ending at $r+k$, retaining updates to future actions and discarding the replay command.
At each live tick, including the transition tick, the RTM uses the latest feedback to update the first three actions in its 24-action window with one Euler step. It sends the first action to the robot and advances the window.

\section{Experiments}
\label{appendix:experiments}

\Cref{appendix:tab:task_protocols} lists the procedure and success criteria for each task.
We first compare RT$^2$C with T-Rex and OpenWAM in \cref{appendix:baseline_comparison}.
We then extend the backbone ablation from Blueberry to Bulb in \cref{appendix:backbone_ablations}.
Finally, we present qualitative results from successful and failed rollouts and analyze the observed failure modes (\cref{appendix:success_cases,appendix:failure_cases}).

\subsection{Comparison with T-Rex and OpenWAM}
\label{appendix:baseline_comparison}
Across the ten tasks in \cref{appendix:tab:additional_baselines}, RT$^2$C achieves a mean success rate of 76.5\%, exceeding T-Rex (37.0\%) by 39.5 percentage points and OpenWAM (46.5\%) by 30.0 percentage points.
RT$^2$C outperforms T-Rex on all ten tasks and OpenWAM on eight. On Blueberry, RT$^2$C succeeds in 95\% of trials, compared with 10\% for each baseline. It matches OpenWAM on Spray (both 100\%) and trails it on Trash (75\% versus 80\%).

\begin{table*}[t]
  \centering
  \caption{Success rates on ten tasks (\%) with more baselines in addition to \cref{tab:exp_main}. Bold marks the best result on each task.}
  \label{appendix:tab:additional_baselines}
  \small
  \setlength{\tabcolsep}{4pt}
  \begin{tabular}{@{}l*{10}{c}@{}}
    \toprule
    & \multicolumn{2}{c}{\shortstack{Gentle grasping\\and holding}} & \multicolumn{3}{c}{Tool squeezing} & \multicolumn{2}{c}{\shortstack{Friction\\and slip}} & \multicolumn{3}{c}{\shortstack{Twisting\\and latching}} \\
    \cmidrule(lr){2-3}\cmidrule(lr){4-6}\cmidrule(lr){7-8}\cmidrule(lr){9-11}
    Method & Blueberry & Bread & Spray & Slime & Dropper & Card & Trash & Bottle cap & Bulb & Container \\
    \midrule
    $\pi_0$~\cite{black2026pi0} & 80 & 75 & \textbf{100} & 30 & \textbf{100} & 55 & 75 & 45 & 85 & 70 \\
    $\pi_{0.5}$~\cite{intelligence2025pi05} & 10 & 90 & \textbf{100} & 10 & 90 & 10 & 80 & 35 & 70 & \textbf{85} \\
    DM0.5~\cite{dexmal2026dm05} & 45 & 65 & \textbf{100} & 10 & 25 & 10 & 10 & 15 & 45 & 80 \\
    MoDE-VLA~\cite{tang2026modevla} & 80 & \textbf{95} & \textbf{100} & \textbf{45} & 95 & \textbf{75} & 60 & \textbf{60} & 80 & 75 \\
    REACT~\cite{xiong2026react} & 30 & 30 & \textbf{100} & 30 & 10 & 50 & \textbf{90} & 40 & 50 & 65 \\
    T-Rex~\cite{niu2026trex} & 10 & 20 & 85 & 25 & 15 & 65 & 60 & 10 & 60 & 20 \\
    OpenWAM~\cite{wang2026openwam} & 10 & 65 & \textbf{100} & 10 & 70 & 70 & 80 & 5 & 40 & 15 \\
    \midrule
    Ours & \textbf{95} & \textbf{95} & \textbf{100} & 30 & \textbf{100} & \textbf{75} & 75 & 30 & \textbf{90} & 75 \\
    \bottomrule
  \end{tabular}
\end{table*}

\begin{table*}[t]
  \centering
  \caption{Backbone ablations on Blueberry and Bulb. Base, +RTC, and +RTC+RTM report success rates (\%). $\Delta$ gives the gain from adding RTM to RTC in percentage points.}
  \label{appendix:tab:bulb_backbones}
  \small
  \setlength{\tabcolsep}{6pt}
  \begin{tabular}{@{}lrrrr@{\hspace{16pt}}rrrr@{}}
    \toprule
    & \multicolumn{4}{c}{Blueberry} & \multicolumn{4}{c}{Bulb} \\
    \cmidrule(lr){2-5}\cmidrule(lr){6-9}
    Backbone & Base & +RTC & +RTC+RTM & $\Delta$ (pp) & Base & +RTC & +RTC+RTM & $\Delta$ (pp) \\
    \midrule
    $\pi_0$     & 80 & 75 & 95 & $+20$ & 85 & 80 & 90 & $+10$ \\
    $\pi_{0.5}$ & 10 & 25 & 35 & $+10$ & 70 & 80 & 80 & $0$ \\
    DM0.5       & 45 & 35 & 40 & $+5$  & 45 & 60 & 90 & $+30$ \\
    \bottomrule
  \end{tabular}
\end{table*}

\subsection{Backbone Ablations}
\label{appendix:backbone_ablations}
\Cref{appendix:tab:bulb_backbones} extends the ablation study on Blueberry in \cref{tab:exp_plugin} to Bulb.
Blueberry tests gentle grasping and holding, while Bulb tests the twisting motion needed to screw a light bulb into its socket.
Adding the RTM raises success by 10 percentage points for $\pi_0$ on Bulb and by 30 points for DM0.5. For $\pi_{0.5}$, success remains at 80\% on Bulb, compared with a 10-point gain on Blueberry.

\subsection{Success Cases}
\label{appendix:success_cases}

\begin{figure*}
    \centering
    \small
    \begin{tabular}{@{}l@{\hspace{3pt}}c@{\hspace{2pt}}c@{\hspace{2pt}}c@{\hspace{2pt}}c@{\hspace{2pt}}c@{}}
        \textbf{Blueberry} &
        \raisebox{-0.5\height}{\includegraphics[width=0.175\textwidth]{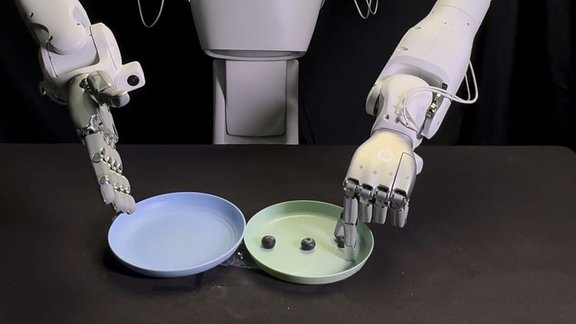}} &
        \raisebox{-0.5\height}{\includegraphics[width=0.175\textwidth]{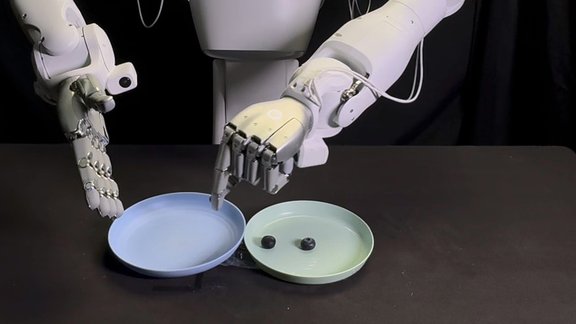}} &
        \raisebox{-0.5\height}{\includegraphics[width=0.175\textwidth]{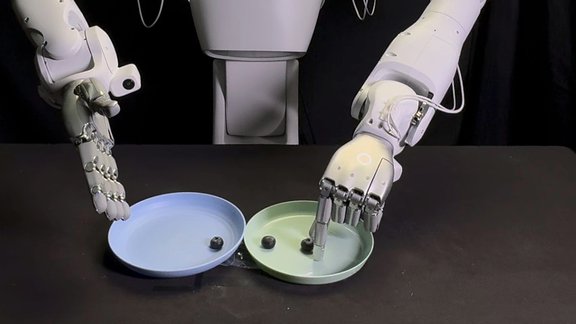}} &
        \raisebox{-0.5\height}{\includegraphics[width=0.175\textwidth]{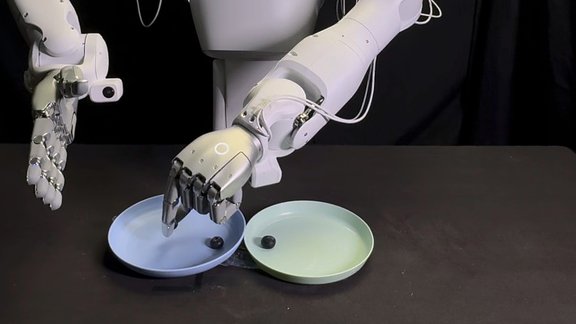}} &
        \raisebox{-0.5\height}{\includegraphics[width=0.175\textwidth]{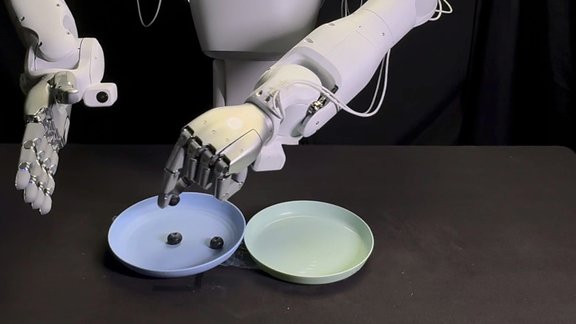}} \\ \noalign{\vskip 3pt}
        \textbf{Bread} &
        \raisebox{-0.5\height}{\includegraphics[width=0.175\textwidth]{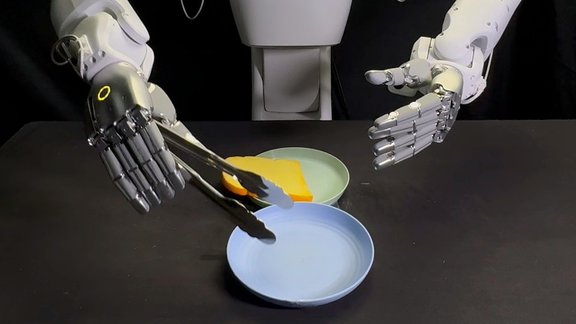}} &
        \raisebox{-0.5\height}{\includegraphics[width=0.175\textwidth]{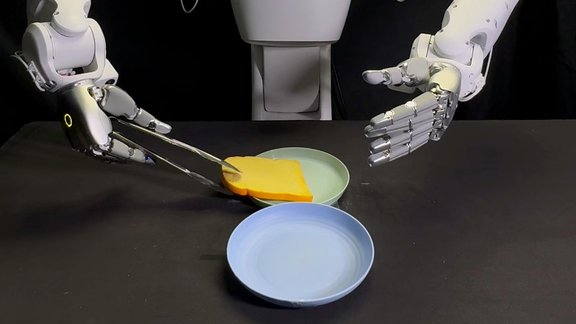}} &
        \raisebox{-0.5\height}{\includegraphics[width=0.175\textwidth]{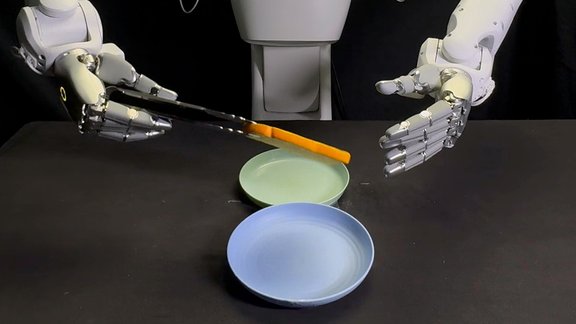}} &
        \raisebox{-0.5\height}{\includegraphics[width=0.175\textwidth]{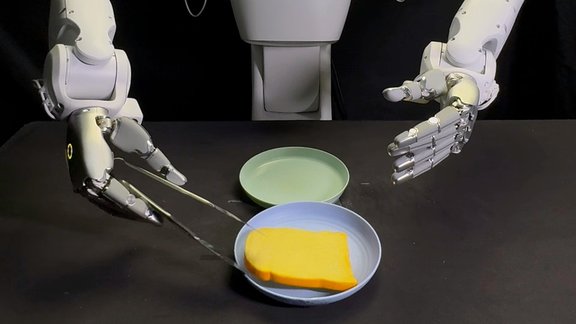}} &
        \raisebox{-0.5\height}{\includegraphics[width=0.175\textwidth]{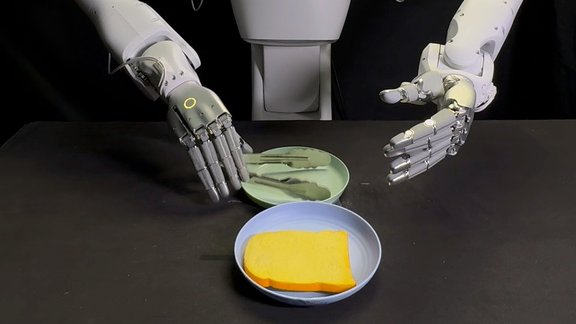}} \\ \noalign{\vskip 3pt}
        \textbf{Spray} &
        \raisebox{-0.5\height}{\includegraphics[width=0.175\textwidth]{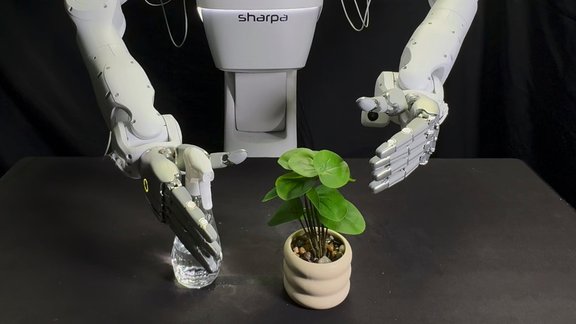}} &
        \raisebox{-0.5\height}{\includegraphics[width=0.175\textwidth]{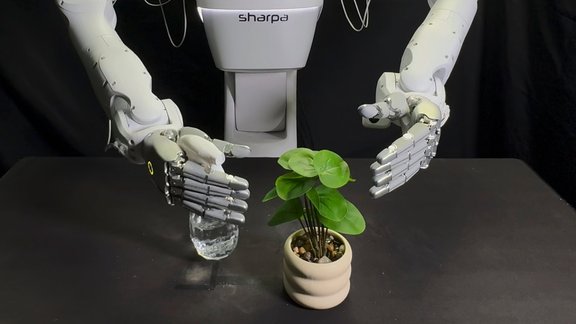}} &
        \raisebox{-0.5\height}{\includegraphics[width=0.175\textwidth]{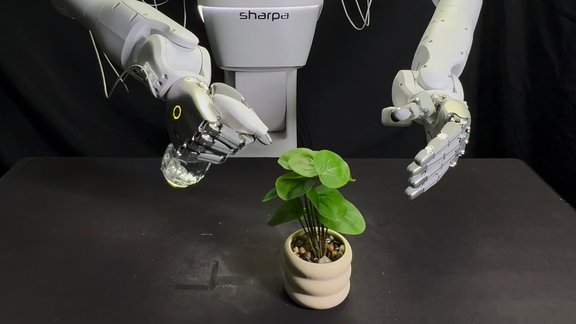}} &
        \raisebox{-0.5\height}{\includegraphics[width=0.175\textwidth]{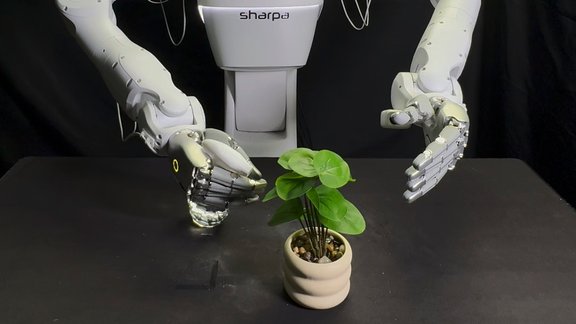}} &
        \raisebox{-0.5\height}{\includegraphics[width=0.175\textwidth]{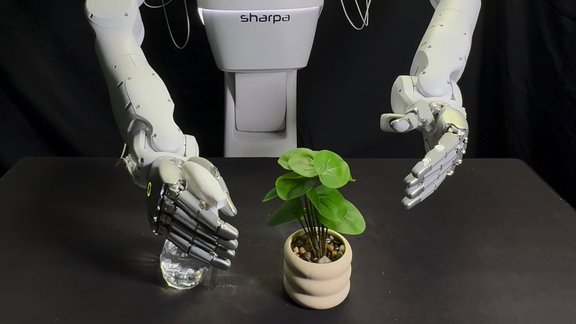}} \\ \noalign{\vskip 3pt}
        \textbf{Slime} &
        \raisebox{-0.5\height}{\includegraphics[width=0.175\textwidth]{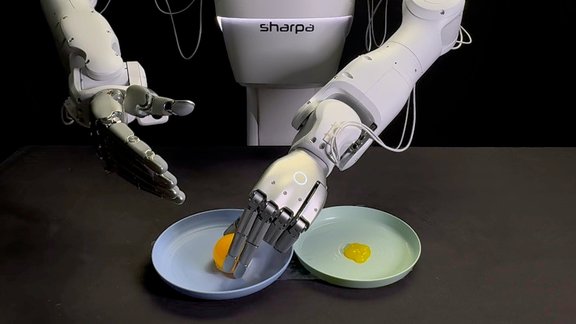}} &
        \raisebox{-0.5\height}{\includegraphics[width=0.175\textwidth]{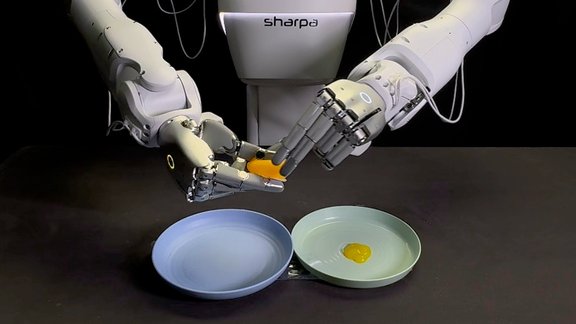}} &
        \raisebox{-0.5\height}{\includegraphics[width=0.175\textwidth]{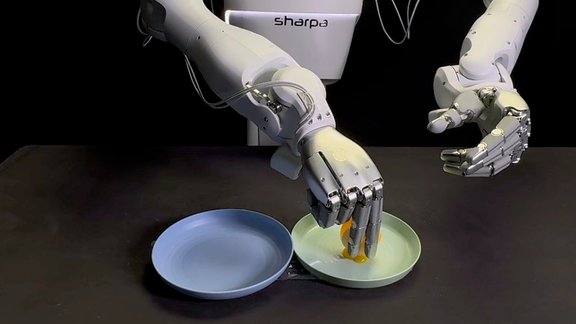}} &
        \raisebox{-0.5\height}{\includegraphics[width=0.175\textwidth]{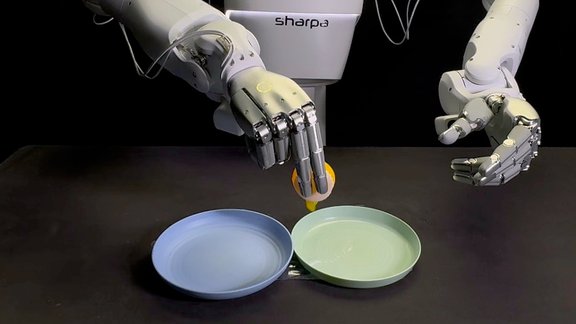}} &
        \raisebox{-0.5\height}{\includegraphics[width=0.175\textwidth]{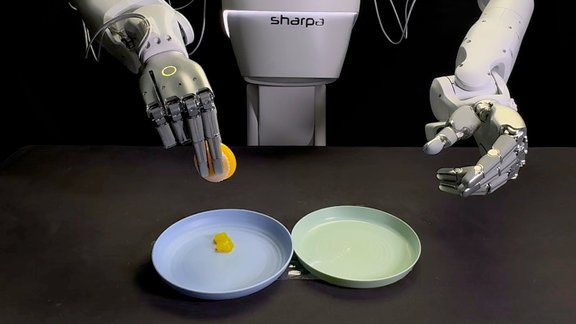}} \\ \noalign{\vskip 3pt}
        \textbf{Dropper} &
        \raisebox{-0.5\height}{\includegraphics[width=0.175\textwidth]{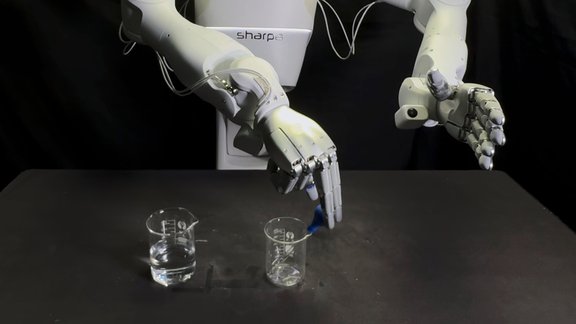}} &
        \raisebox{-0.5\height}{\includegraphics[width=0.175\textwidth]{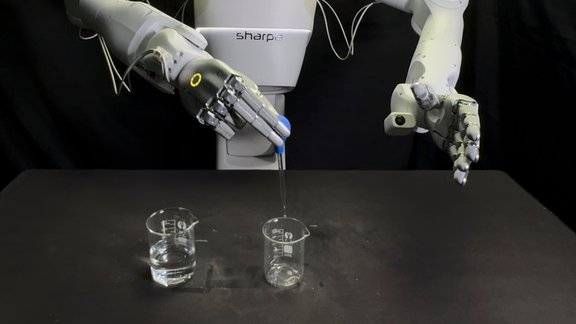}} &
        \raisebox{-0.5\height}{\includegraphics[width=0.175\textwidth]{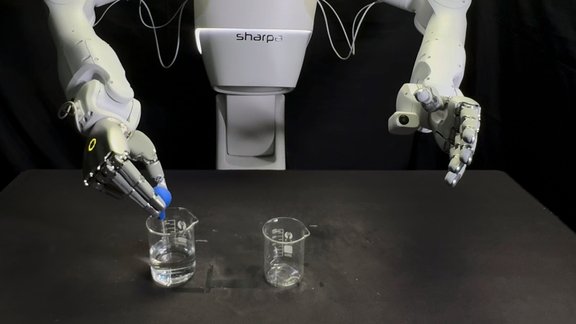}} &
        \raisebox{-0.5\height}{\includegraphics[width=0.175\textwidth]{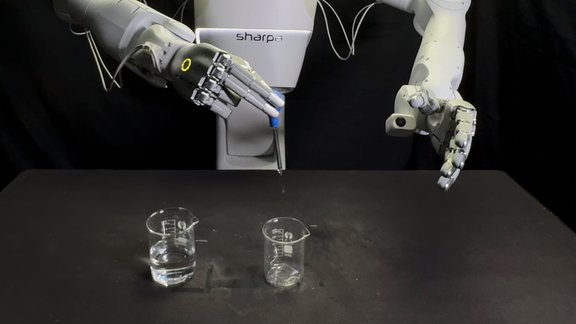}} &
        \raisebox{-0.5\height}{\includegraphics[width=0.175\textwidth]{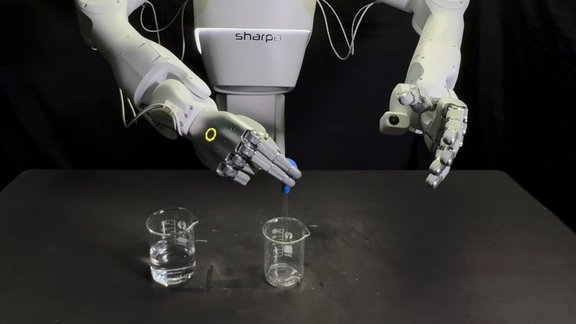}} \\ \noalign{\vskip 3pt}
        \textbf{Card} &
        \raisebox{-0.5\height}{\includegraphics[width=0.175\textwidth]{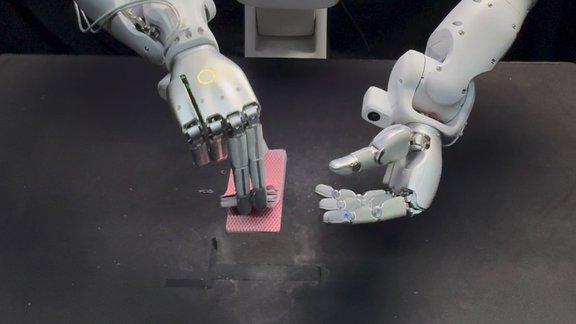}} &
        \raisebox{-0.5\height}{\includegraphics[width=0.175\textwidth]{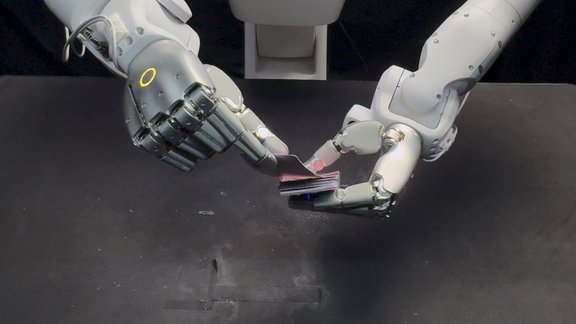}} &
        \raisebox{-0.5\height}{\includegraphics[width=0.175\textwidth]{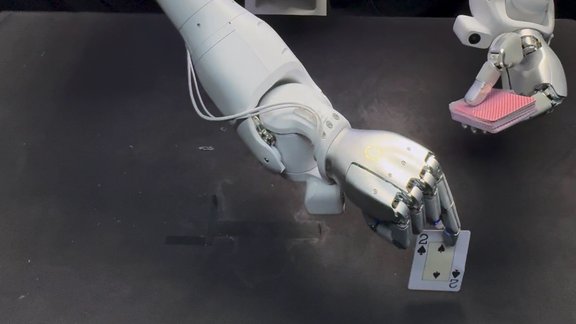}} &
        \raisebox{-0.5\height}{\includegraphics[width=0.175\textwidth]{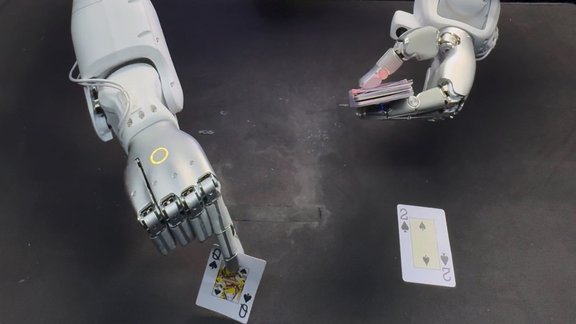}} &
        \raisebox{-0.5\height}{\includegraphics[width=0.175\textwidth]{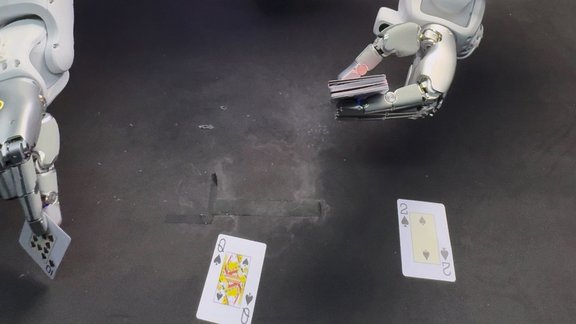}} \\ \noalign{\vskip 3pt}
        \textbf{Trash} &
        \raisebox{-0.5\height}{\includegraphics[width=0.175\textwidth]{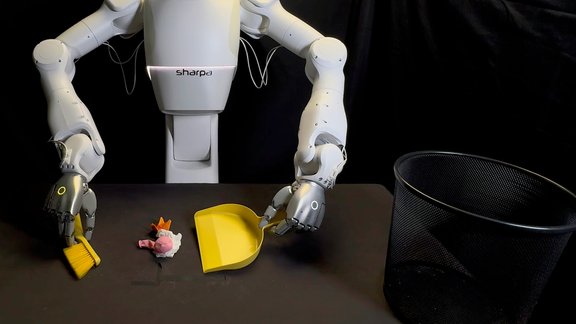}} &
        \raisebox{-0.5\height}{\includegraphics[width=0.175\textwidth]{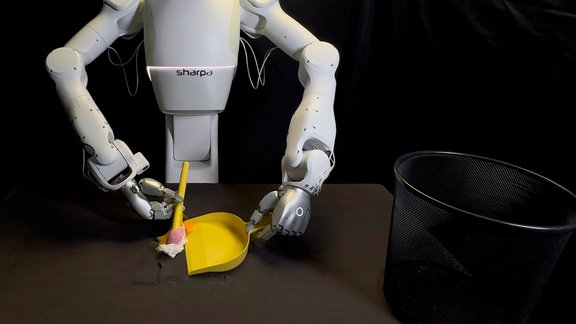}} &
        \raisebox{-0.5\height}{\includegraphics[width=0.175\textwidth]{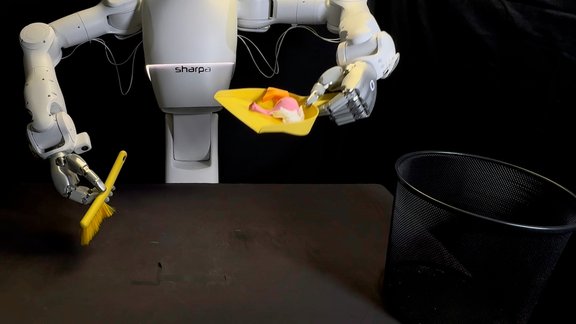}} &
        \raisebox{-0.5\height}{\includegraphics[width=0.175\textwidth]{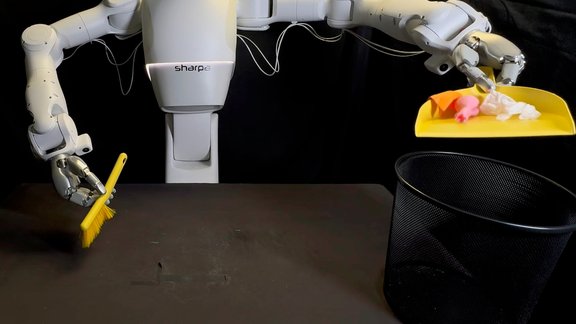}} &
        \raisebox{-0.5\height}{\includegraphics[width=0.175\textwidth]{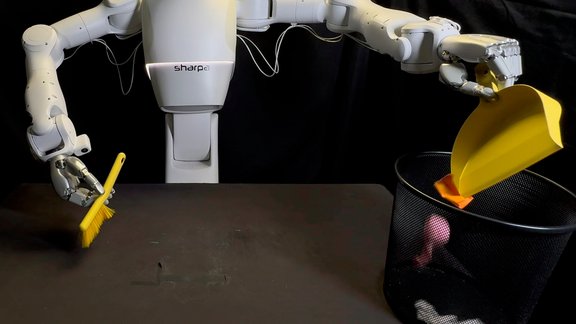}} \\ \noalign{\vskip 3pt}
        \textbf{Bottle cap} &
        \raisebox{-0.5\height}{\includegraphics[width=0.175\textwidth]{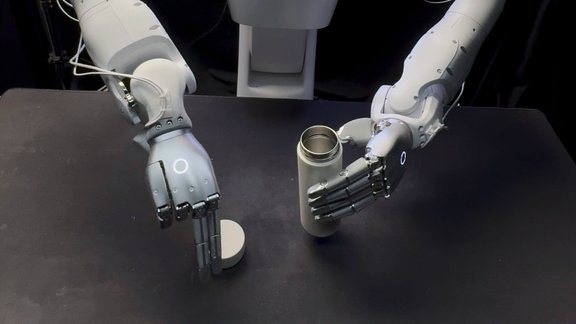}} &
        \raisebox{-0.5\height}{\includegraphics[width=0.175\textwidth]{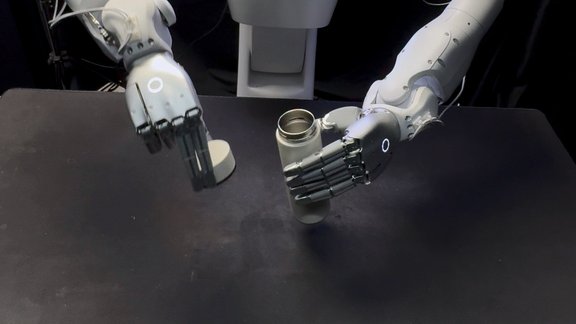}} &
        \raisebox{-0.5\height}{\includegraphics[width=0.175\textwidth]{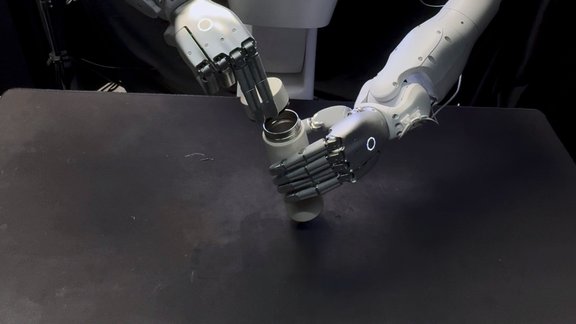}} &
        \raisebox{-0.5\height}{\includegraphics[width=0.175\textwidth]{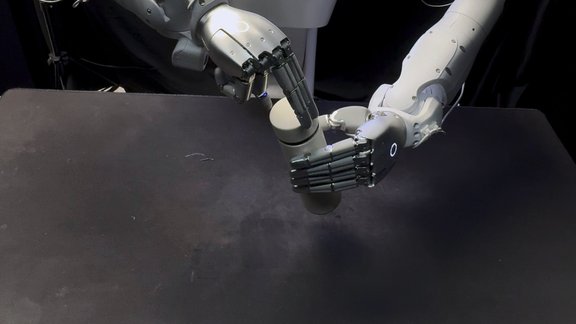}} &
        \raisebox{-0.5\height}{\includegraphics[width=0.175\textwidth]{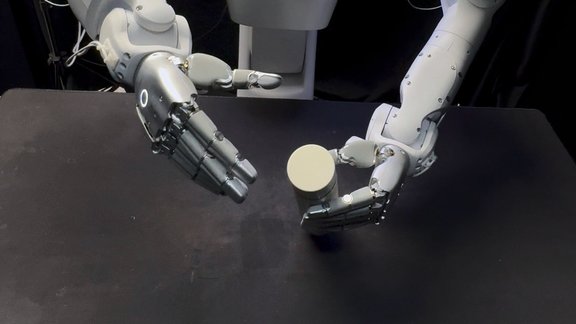}} \\ \noalign{\vskip 3pt}
        \textbf{Bulb} &
        \raisebox{-0.5\height}{\includegraphics[width=0.175\textwidth]{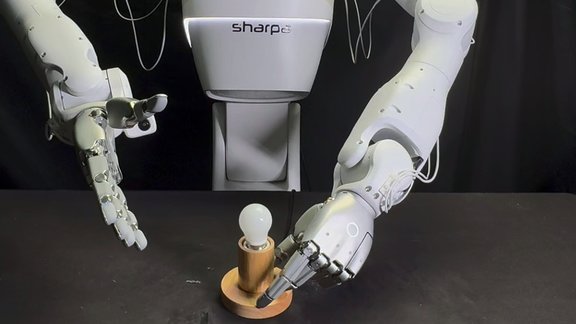}} &
        \raisebox{-0.5\height}{\includegraphics[width=0.175\textwidth]{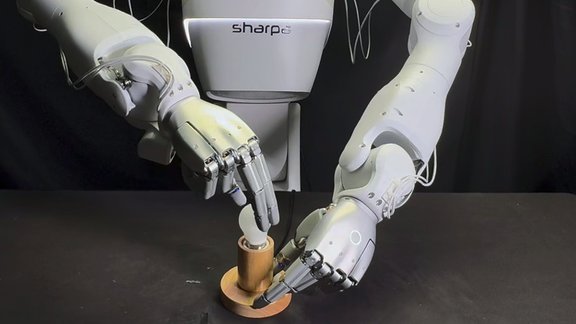}} &
        \raisebox{-0.5\height}{\includegraphics[width=0.175\textwidth]{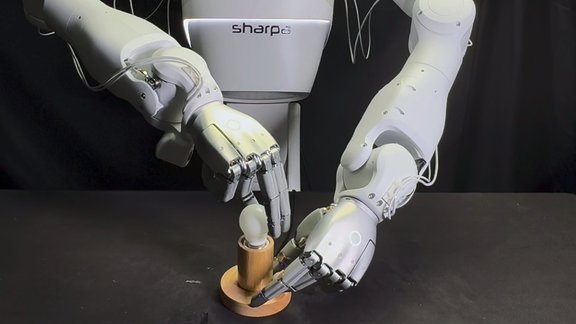}} &
        \raisebox{-0.5\height}{\includegraphics[width=0.175\textwidth]{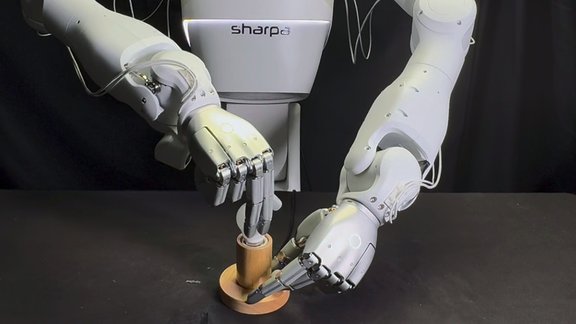}} &
        \raisebox{-0.5\height}{\includegraphics[width=0.175\textwidth]{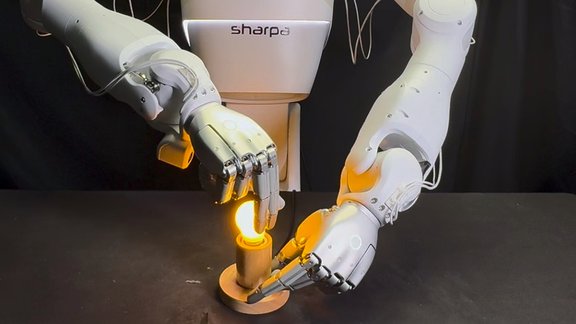}} \\ \noalign{\vskip 3pt}
        \textbf{Container} &
        \raisebox{-0.5\height}{\includegraphics[width=0.175\textwidth]{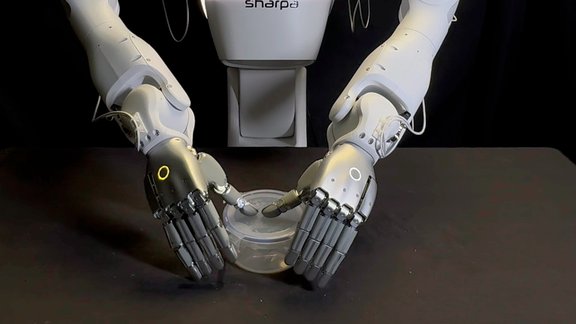}} &
        \raisebox{-0.5\height}{\includegraphics[width=0.175\textwidth]{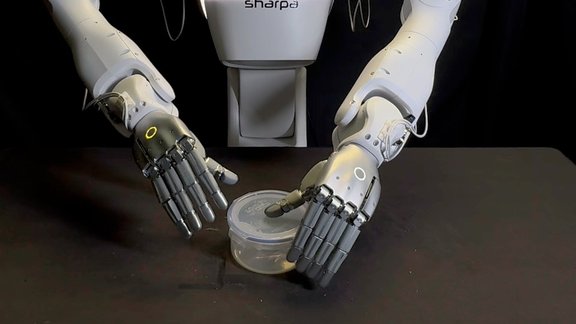}} &
        \raisebox{-0.5\height}{\includegraphics[width=0.175\textwidth]{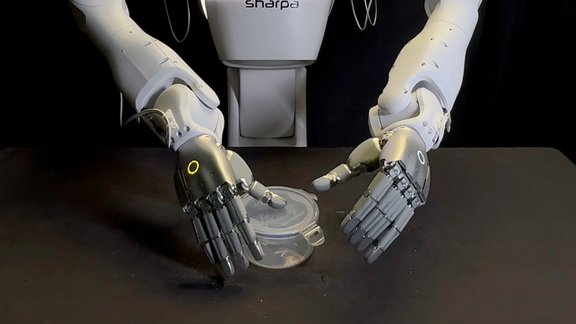}} &
        \raisebox{-0.5\height}{\includegraphics[width=0.175\textwidth]{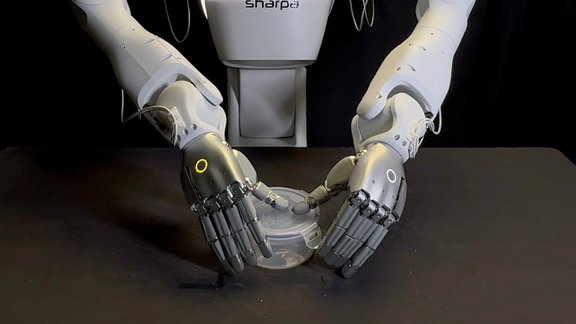}} &
        \raisebox{-0.5\height}{\includegraphics[width=0.175\textwidth]{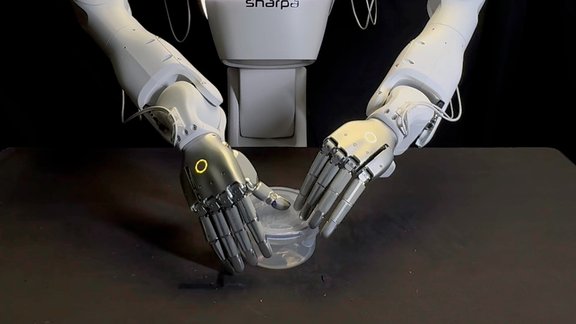}}
    \end{tabular}
    \caption{\textbf{Success cases from a third-person view.} Five frames are shown for each of the ten tasks, complementing the head-camera views in \cref{fig:exp_tasks}. Frames are cropped to highlight hand--object interactions. Rows follow the task order in \cref{appendix:tab:task_protocols}, and frames progress chronologically from left to right.}
    \label{appendix:fig:success_cases}
\end{figure*}

\Cref{appendix:fig:success_cases} presents successful executions of all ten tasks from an external, third-person camera. These views complement the head-camera views in \cref{fig:exp_tasks} by showing hand--object interactions and the relevant workspace. Each row contains five chronologically ordered frames illustrating key stages of one successful rollout.

\subsection{Failure Cases}
\label{appendix:failure_cases}

\begin{figure*}[p]
    \centering
    \captionsetup{singlelinecheck=false,skip=4pt}
    \captionsetup[subfigure]{hypcap=true,font=small,labelfont=bf,justification=raggedright,singlelinecheck=false,skip=2pt}
    \begin{subfigure}{\linewidth}
        \includegraphics[width=0.329\linewidth]{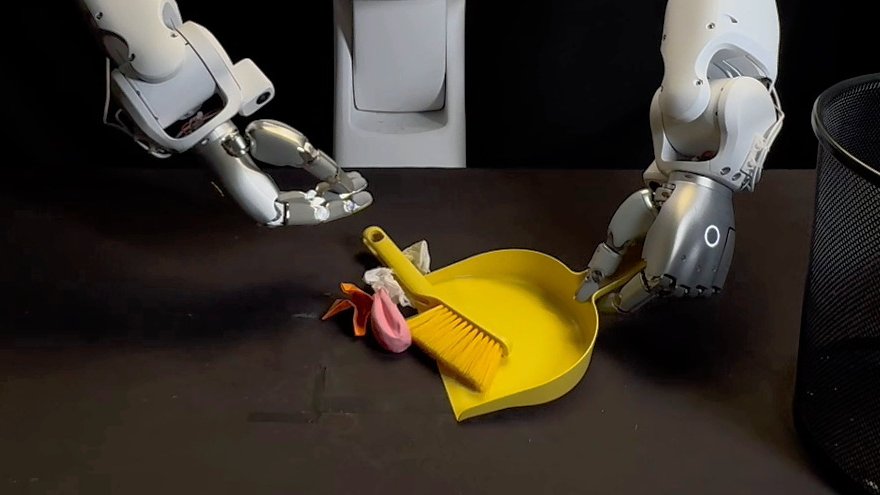}\hfill
        \includegraphics[width=0.329\linewidth]{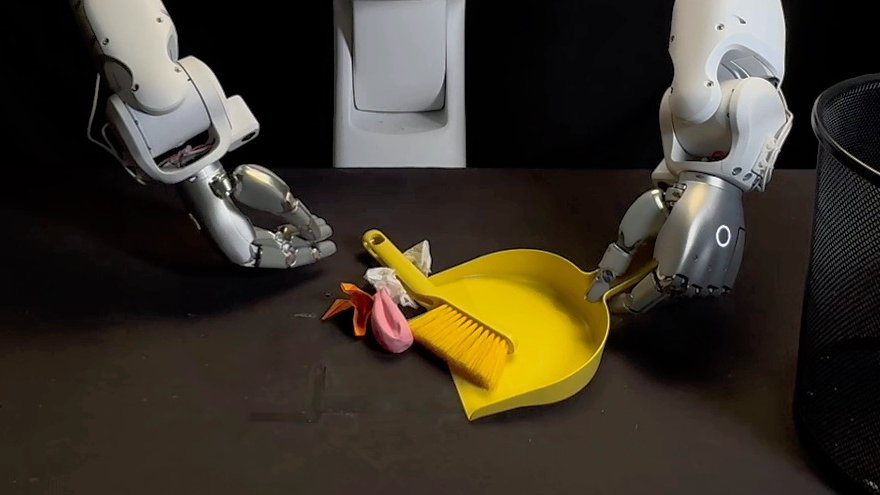}\hfill
        \includegraphics[width=0.329\linewidth]{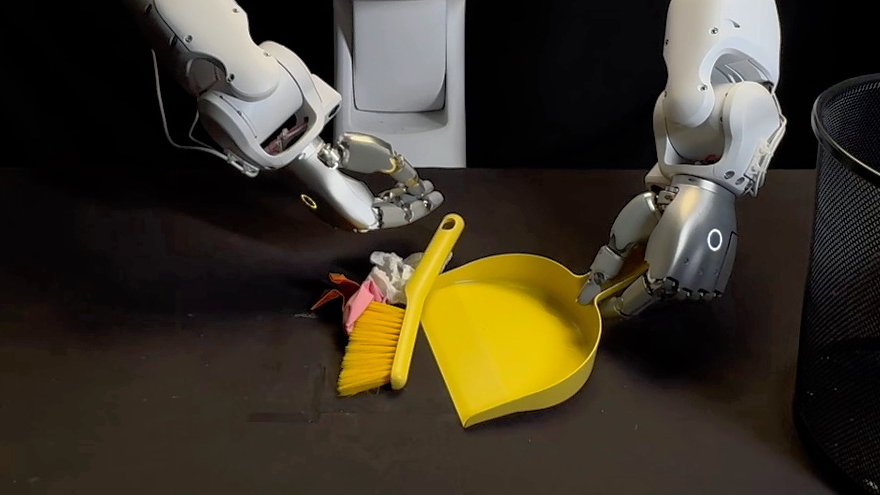}
        \caption{\textbf{Failure case on Trash.} The broom slips out of the right hand and rests against the dustpan, interrupting the sweep. The hand fails to recover a secure grasp in the displayed sequence.}
        \label{appendix:fig:failure_trash}
    \end{subfigure}\par\addvspace{2pt}
    \begin{subfigure}{\linewidth}
        \includegraphics[width=0.329\linewidth]{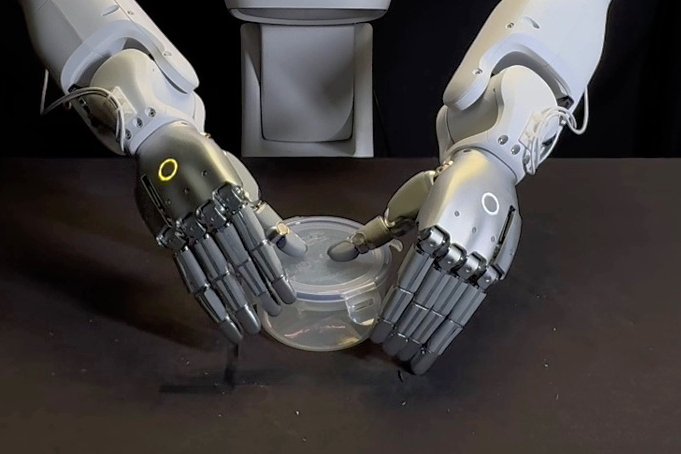}\hfill
        \includegraphics[width=0.329\linewidth]{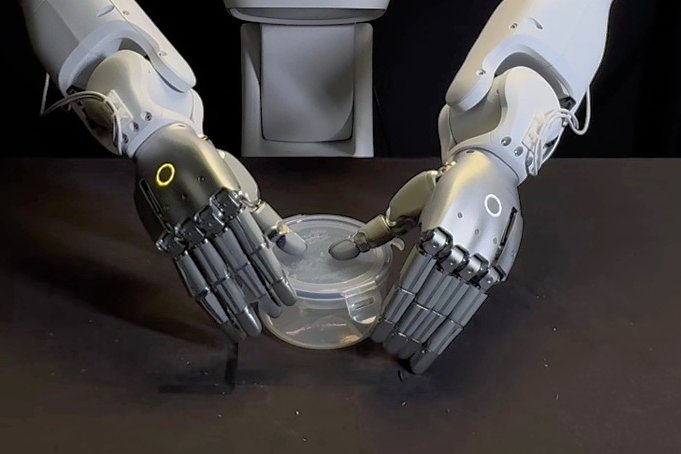}\hfill
        \includegraphics[width=0.329\linewidth]{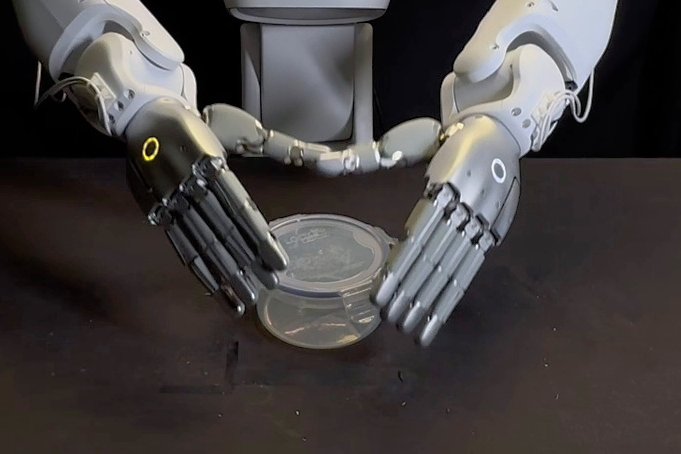}
        \caption{\textbf{Failure case on Container.} The robot presses the latch with the middle phalanx (the second finger segment) instead of the fingertip. Because the latch contact lies outside the fingertip tactile sensing area, the policy does not use tactile feedback to determine whether the latch has successfully closed. The hands begin to withdraw while the latch remains unfastened.}
        \label{appendix:fig:failure_container}
    \end{subfigure}\par\addvspace{2pt}
    \begin{subfigure}{\linewidth}
        \includegraphics[width=0.329\linewidth]{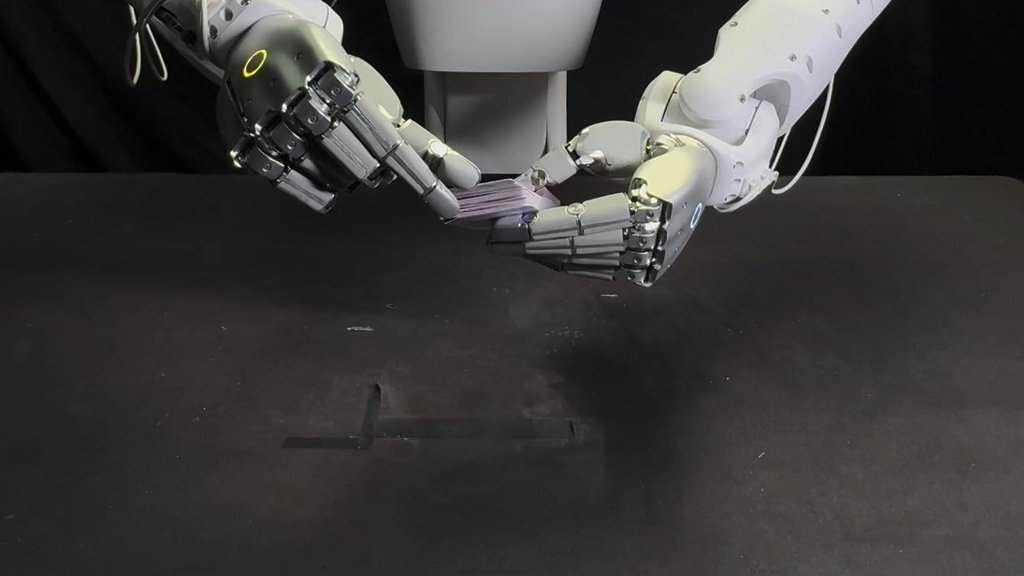}\hfill
        \includegraphics[width=0.329\linewidth]{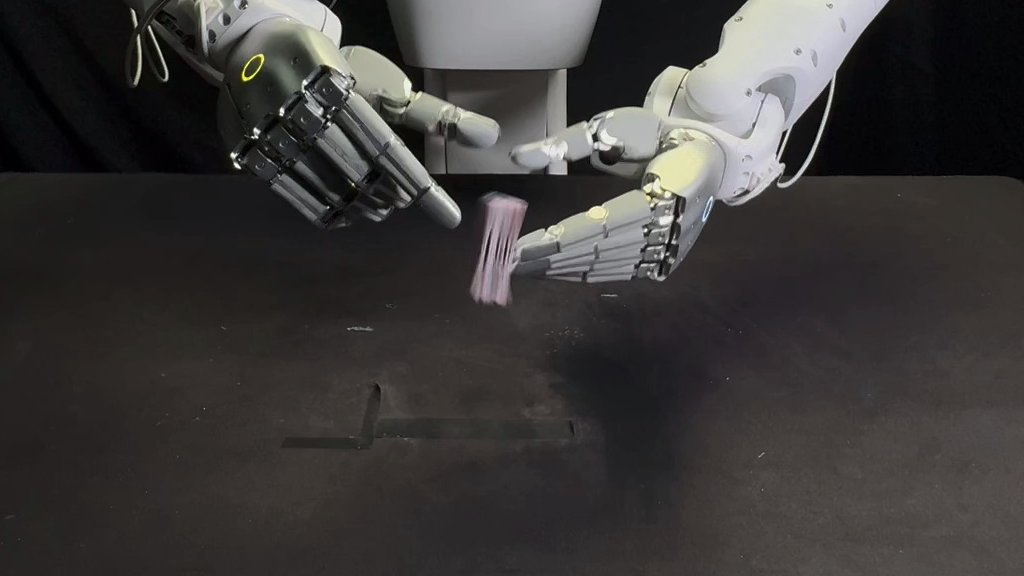}\hfill
        \includegraphics[width=0.329\linewidth]{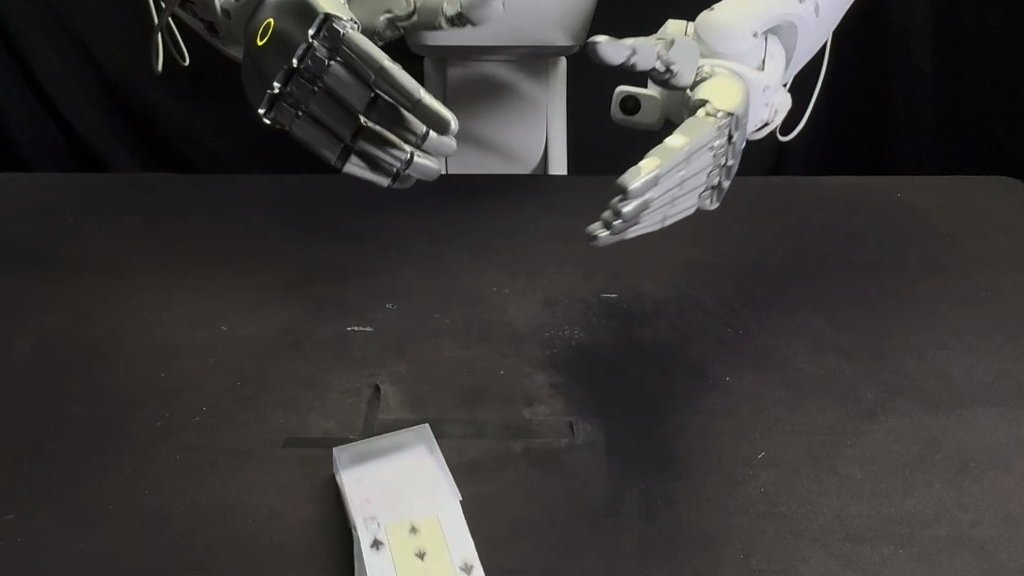}
        \caption{\textbf{Failure case on Card.} The card stack slips from between the hands and falls onto the table.}
        \label{appendix:fig:failure_card}
    \end{subfigure}\par\addvspace{2pt}
    \begin{subfigure}{\linewidth}
        \includegraphics[width=0.329\linewidth]{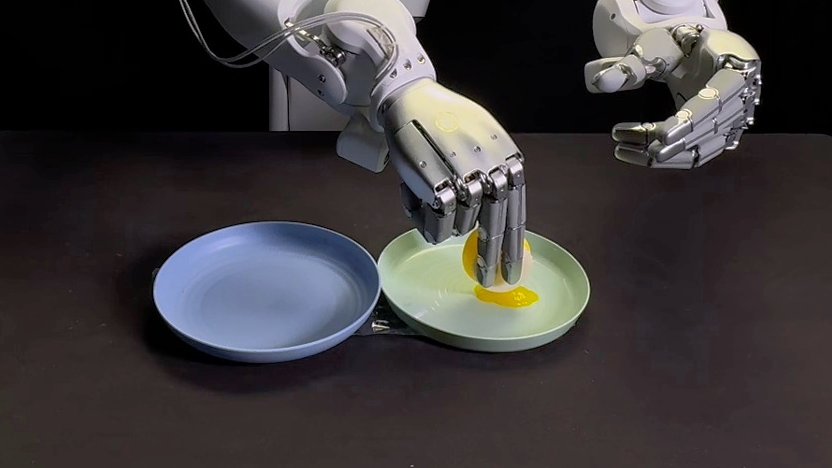}\hfill
        \includegraphics[width=0.329\linewidth]{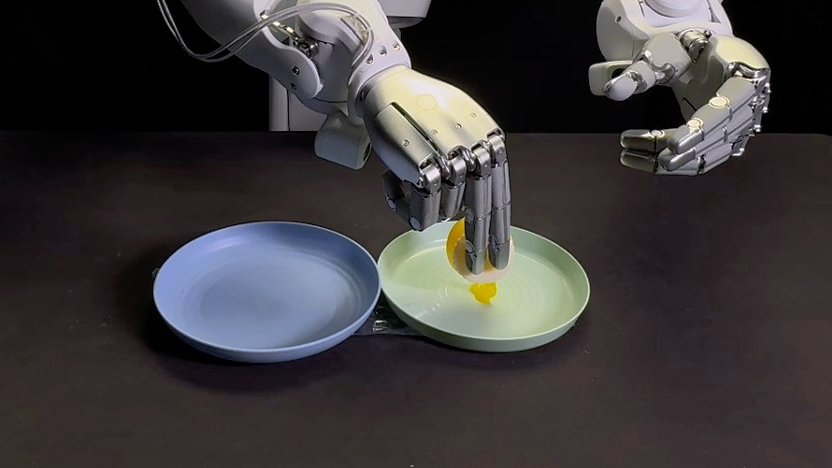}\hfill
        \includegraphics[width=0.329\linewidth]{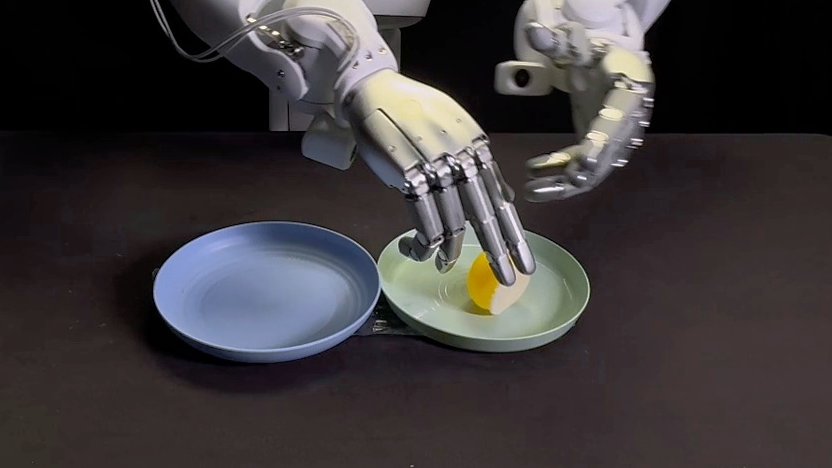}
        \caption{\textbf{Failure case on Slime.} The policy gets stuck at one location, repeatedly repositioning and squeezing the slime but failing to generate sufficient suction to lift it. The slime remains on the green source plate, and the blue target plate remains empty.}
        \label{appendix:fig:failure_slime}
    \end{subfigure}\par\addvspace{2pt}
    \begin{subfigure}{\linewidth}
        \includegraphics[width=0.329\linewidth]{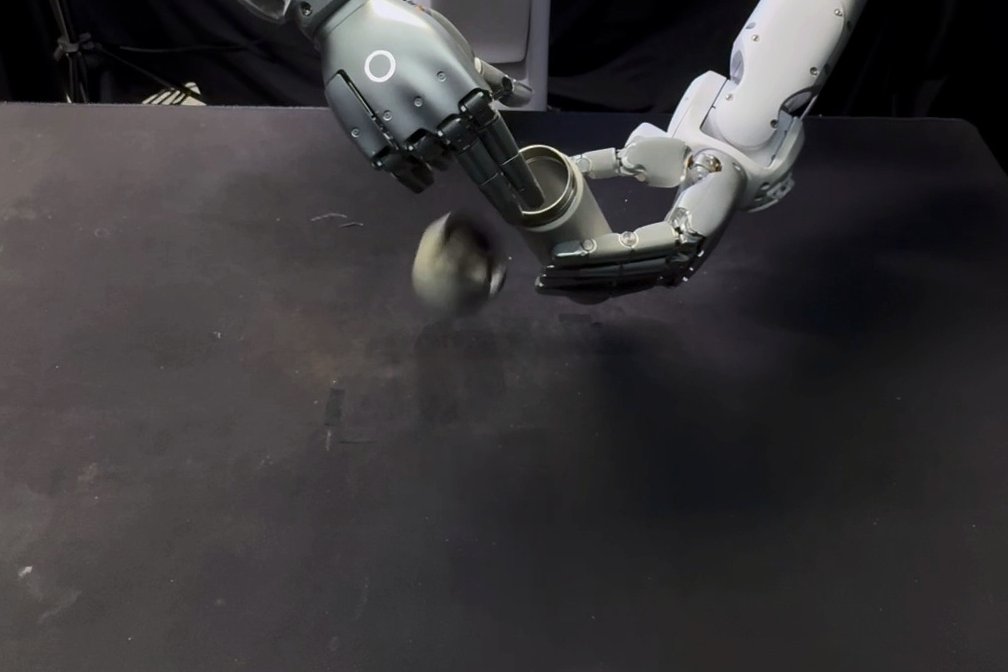}\hfill
        \includegraphics[width=0.329\linewidth]{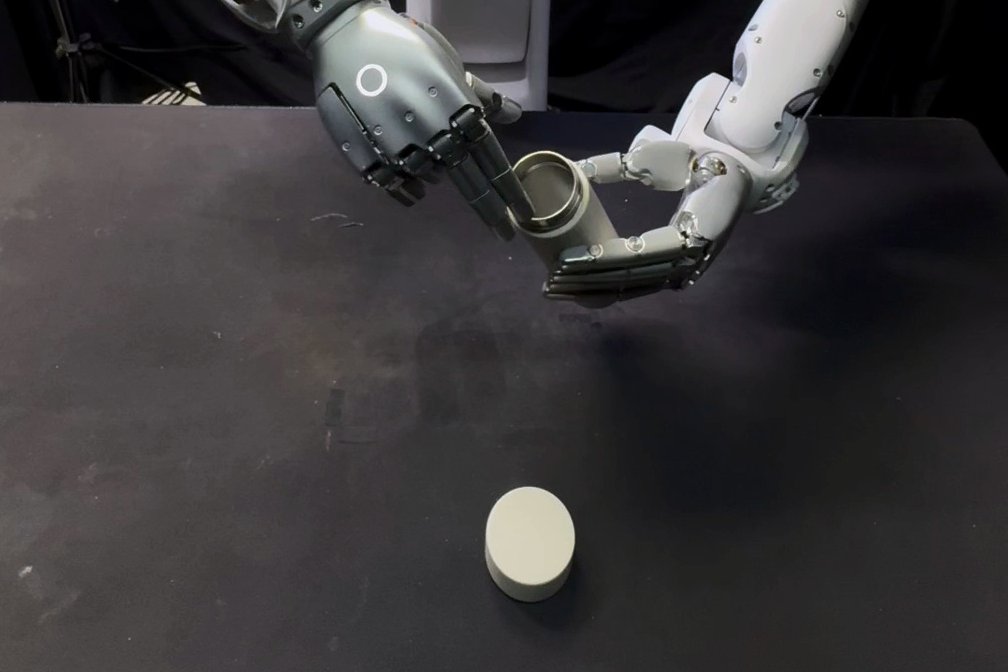}\hfill
        \includegraphics[width=0.329\linewidth]{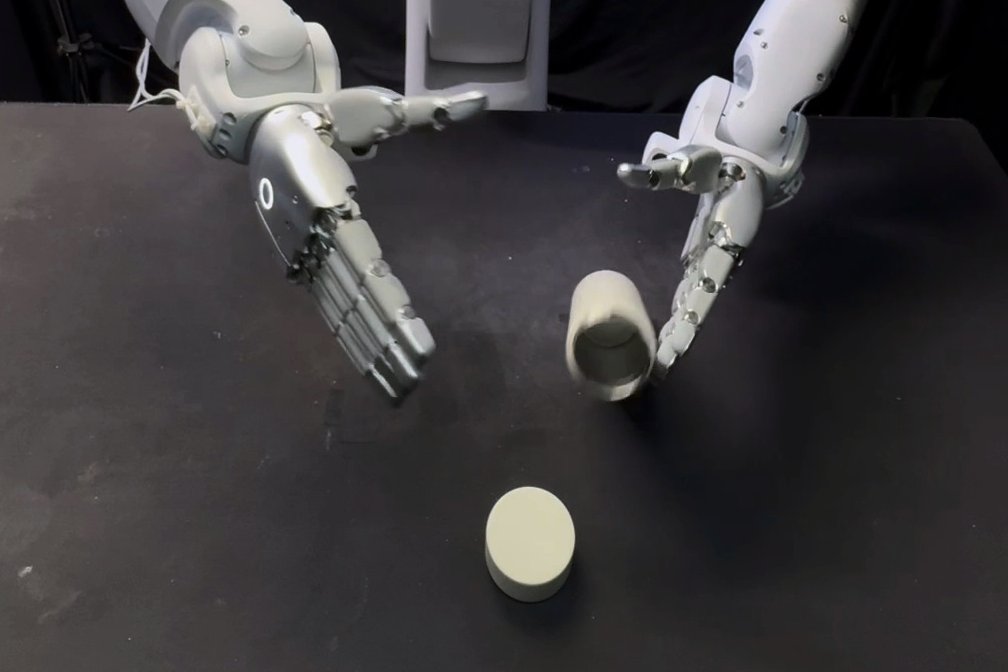}
        \caption{\textbf{Failure case on Bottle cap.} The cap slips off the bottle opening and lands on the table. The bottle remains uncapped and begins to tip over when the holding hand releases it.}
        \label{appendix:fig:failure_bottle_cap}
    \end{subfigure}
    \caption{\textbf{Representative failure cases.} Each row shows three frames from one rollout, ordered chronologically from left to right.}
    \label{appendix:fig:failure_cases}
\end{figure*}

\Crefrange{appendix:fig:failure_trash}{appendix:fig:failure_bottle_cap} show representative failure sequences on five tasks, with frames ordered chronologically from left to right. On Trash, Card, and Bottle cap, the robot loses control of the manipulated object: the broom slips out of the hand, the card stack falls between the hands, and the bottle cap drops before it is secured. These examples show that continuous feedback does not always prevent slippage during grasping and manipulation.

The Container failure (\cref{appendix:fig:failure_container}) illustrates a limitation of fingertip sensing. The robot contacts the latch with the middle phalanx rather than the fingertip, leaving the relevant contact outside the tactile sensing area. As a result, the policy does not use tactile feedback to verify whether the latch has closed, and the hands withdraw with the latch still unfastened. This case highlights that effective tactile feedback depends on establishing contact with the instrumented part of the hand.

On Slime (\cref{appendix:fig:failure_slime}), the policy gets stuck at one location, repeatedly repositioning and squeezing the slime without generating sufficient suction to lift it. The repeated adjustments fail to produce a successful pickup, leaving the slime on the source plate. As discussed in \cref{sec:exp_main}, we suspect that the RTM's per-step corrections can become counterproductive in this relatively tolerant task, which does not require fine tactile feedback. This motivates further study of when tactile corrections should be applied.

\end{document}